\documentclass[10pt,journal,compsoc]{IEEEtran}

\ifCLASSOPTIONcompsoc
  \usepackage[nocompress]{cite}
\else
  \usepackage{cite}
\fi

\ifCLASSINFOpdf
   \usepackage[pdftex]{graphicx}
\else
   \usepackage[dvips]{graphicx}
\fi

\usepackage[cmex10]{amsmath}
\usepackage{amssymb}
\usepackage[linesnumbered,lined,commentsnumbered,ruled]{algorithm2e}
\usepackage{eqparbox}
\usepackage{stfloats}
\usepackage{url}
\usepackage{rotating}
\usepackage{subfig}
\usepackage{epsfig}
\usepackage{booktabs,tabularx}
\usepackage{multirow}
\usepackage[table]{xcolor}
\usepackage{float}
\usepackage{arydshln}
\usepackage{dsfont}
\usepackage{pifont}
\usepackage[para,online,flushleft]{threeparttable}
\usepackage[pagebackref=false,breaklinks=true,letterpaper=true,colorlinks,bookmarks=false]{hyperref}

\newcommand{\cmark}{\ding{51}}%

\begin{document}

\title{Human Activity Recognition Using Robust\\ Adaptive Privileged Probabilistic Learning}
\author{Michalis~Vrigkas,
        Evangelos~Kazakos,
        Christophoros~Nikou,
        and~Ioannis~A.~Kakadiaris
}

\IEEEtitleabstractindextext{%
\begin{abstract}
In this work, a novel method based on the learning using
privileged information (LUPI) paradigm for recognizing complex
human activities is proposed that handles missing information
during testing. We present a supervised probabilistic approach
that integrates LUPI into a hidden conditional random field
(HCRF) model. The proposed model is called HCRF+ and may be
trained using both maximum likelihood and maximum margin
approaches. It employs a self-training technique for automatic
estimation of the regularization parameters of the objective
functions. Moreover, the method provides robustness to outliers
(such as noise or missing data) by modeling the conditional
distribution of the privileged information by a Student's
\textit{t}-density function, which is naturally integrated into
the HCRF+ framework. Different forms of privileged information
were investigated. The proposed method was evaluated using four
challenging publicly available datasets and the experimental
results demonstrate its effectiveness with respect to
the-state-of-the-art in the LUPI framework using both
hand-crafted features and features extracted from a
convolutional neural network.
\end{abstract}

\begin{IEEEkeywords}
Hidden conditional random fields, learning using privileged
information, human activity recognition
\end{IEEEkeywords}}

\maketitle

\section{Introduction}
\label{sec:introduction} 

Recent advances in computer vision such as video
surveillance and human-machine interactions
\cite{Cohen04,Smeulders13} rely on machine learning techniques
trained on large scale human annotated datasets. However, training
data may not always be available during testing and learning using
privileged information (LUPI) \cite{Vapnik09,Lopez_PazBSV16} has been used to
overcome this problem. The insight of privileged information is that
one may have access to additional information about the training
samples, which is not available during testing.

Consequently, classification models may often suffer from ``structure
imbalance'' between training and testing data, which may be
represented by the LUPI paradigm. 
The LUPI technique simulates a real-life learning condition,
when a student learns from his/her teacher, where the latter
provides the student with additional knowledge, comments,
explanations, or rewards in class. Subsequently, the student
should be able to face any problem related to what he/she has
learned without the help of the teacher. Taking advantage of
this learning model, the LUPI framework has also been used in
several machine learning applications such as
boosting \cite{JChen12} and clustering \cite{Feyereisl12}.

The problem of human activity understanding using privileged
knowledge is on its own a very challenging task. Since
privileged information is only available during training, one
should combine both regular and privileged information into a
unified classifier to predict the true class label. 
However, it is quite difficult to identify the most useful
information to be used as privileged as the lack of informative
data or the presence of misleading information may influence
the performance of the model. 

\begin{figure}[!t]
\centering
\includegraphics[width=0.9\columnwidth, clip=true]{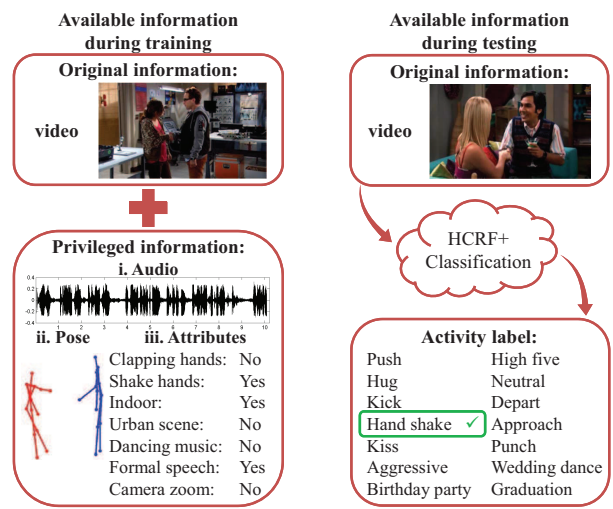}
\caption{Robust learning using privileged information. Given a set of training
examples and a set of additional information about the training samples
(left), our system can successfully recognize the class label of the
underlying activity without having access to the additional information
during testing (right). We explore three different forms of privileged
information (e.g., audio signals, human poses, and attributes) by
modeling them with a Student's \textit{t}-distribution and incorporating
them into the HCRF+ model.}
\label{fig:figure1}
\end{figure}

We address these issues by presenting a new probabilistic
approach, based on hidden conditional random fields (HCRFs)
\cite{Quattoni07}, called HCRF+. The proposed method is able to learn human
activities by exploiting additional information about the input
data, that may reflect on natural or auxiliary properties about
classes and members of the classes of the training data (Fig. \ref{fig:figure1}) and it is used for training purposes
only but not for predicting the true classes (where, in
general, this information is missing).

In particular, the proposed HCRF+ method differentiates from
previous approaches \cite{vrigkas2017inferring}, which may also use the LUPI paradigm, by
incorporating privileged information in a supervised
probabilistic manner, which facilitates the training process by
learning the conditional probability distribution between human
activities and observations. We also introduce a novel
technique for automatic estimation of the optimal
regularization parameters for the learning process. The method
is adaptive as the regularization parameters are computed from
the training data through a self-training procedure. It is
worth noting that the proposed methodology is not limited to
the use of a specific form of privileged information, but it is
general and may handle any form of additional data.

Moreover, our method can efficiently manage dissimilarities in
input data, which may correspond to noise, missing data, or
outliers, using a Student's \textit{t}-distribution to model
the conditional probability of the privileged information. Such
dissimilarities may harm the classification accuracy and lead
to excessive sensitivity when input data is insufficient or contains
large intra-class variations. In particular, the use of
Student's \textit{t}-distribution is justified by the property
that it has heavier tails than a standard Gaussian
distribution, thus providing robustness to outliers
\cite{Peel00}.

The main contributions of our work can be summarized in the
following points. A human activity recognition method is
proposed, which exploits privileged information in a
probabilistic manner by introducing a novel classification
scheme based on HCRFs to deal with missing or incomplete data
during testing. Both maximum likelihood and maximum margin
approaches are incorporated into the proposed HCRF+ model.
Moreover, a novel technique for adaptive estimation of the
regularization term during the learning process is introduced
by incorporating both privileged and regular data. Finally,
contrary to previous methods, which may be sensitive to
outlying data measurements, a robust framework for recognizing
human activities is intergraded by employing a Student's
\textit{t}-distribution to attain robustness against outliers.

The remainder of the paper is organized as follows: in Section
\ref{sec:realatedwork}, a review of the related work is presented.
Section \ref{sec:approach} presents the proposed HCRF+ approach
including the maximum likelihood and maximum margin approaches for
learning the model's parameters and the automatic estimation of the
regularization terms. In Section \ref{sec:experiments}, experimental results are
reported, and a discussion about the performance of the
proposed approach is offered in Section \ref{sec:discussion}.
Finally, conclusions are drawn in Section
\ref{sec:conclusions}.

\section{Related Work}
\label{sec:realatedwork}

A major family of methods relies on learning human activities
by building visual models and assigning activity roles to
people associated with an event \cite{Ramanathan13a,WangS13a}.
In recent years, there has been an increased focus
on the combination of different kinds of modalities, such as
visual and audio information, for activity classification
\cite{SongMD12,MVrigkasTAC17}.
A shared representation of human poses and visual information
has also been explored \cite{Ferrari09,Ykiwon12}. However, the
effectiveness of such methods is limited by tracking
inaccuracies in human poses and complex backgrounds. To this
end, Cherian \emph{et al.} \cite{CherianMAS14} explored several
kinematic and part-occlusion constraints for decomposing human
poses into separate limbs to localize the human body. Eweiwi
\emph{et al.} \cite{EweiwiCBG14} reduced the required amount of
pose data using a fixed length vector of more informative
motion features for each skeletal point. 

Special focus has also been given in recognizing human
activities from movies or TV shows by exploiting scene contexts
to localize activities and understand human interactions
\cite{Perez12,HoaiZ14}. Ramanathan \emph{et al.}
\cite{RamanathanLL13} improved the recognition accuracy of such
complex videos by relating textual descriptions and visual
context to a unified framework. Guadarrama \emph{et al.}
\cite{GuadarramaKMVMDS13} proposed an alternative to the
previous approach that takes a video clip as input and
generates short textual descriptions, which may correspond to
an activity label that is unseen during training. However, natural
video sequences may contain irrelevant
scenes or scenes with multiple actions. 
Shao \emph{et al.} \cite{Shao15} mixed appearance and motion
features using multi-task deep learning for recognizing group
activities in crowded scenes collected from the web. 
Mar\'{\i}n-Jim\'{e}nez \emph{et al.} \cite{MarinJemenez14} used a bag
of visual-audio words scheme along with late fusion for recognizing
human interactions in TV shows. Even though their method performs
well in recognizing human interaction, the lack of an intrinsic
audio-visual relationship estimation limits the recognition problem.

Intermediate semantic features representation for recognizing
unseen actions during training has been extensively studied
\cite{Wang2010,JLiu11,FuHXG12}. These intermediate features are
learned during training and enable parameter sharing between
classes by capturing the correlations between frequently
occurring low-level features \cite{AkataPHS13}.

Recent methods that exploited deep neural networks have
demonstrated remarkable results in large-scale datasets
\cite{Carreira_2017_CVPR}. Donahue \emph{et al.}
\cite{lrcn2014} proposed a recurrent convolutional
architecture, where long short-term memory (LSTM) 
networks \cite{HochreiterS97} are connected to
convolutional neural networks (CNNs) that can be jointly
trained to simultaneously learn spatio-temporal dynamics. Wang
\emph{et al.} \cite{Wang_2015_CVPR} presented a new video
representation that employs CNNs to learn multi-scale
convolutional feature maps and introduced the strategies of
trajectory-constrained sampling and pooling to encode deep
features into informative descriptors. Tran \emph{et al.}
\cite{tran2015learning} introduced a 3D ConvNet architecture
that learns spatio-temporal features using 3D convolutions. A
novel video representation, that can summarize a video into a
single image by applying rank pooling on the raw image pixels,
was proposed by Bilen \emph{et al.} \cite{Bilen_2016_CVPR}.
Feichtenhofer \emph{et al.} \cite{Feichtenhofer_2016_CVPR}
introduced a novel architecture for two stream ConvNets and
studied different ways for spatio-temporal fusion of the
ConvNet towers. Zhu \emph{et al.} \cite{Zhu_2016_CVPR} argued
that videos contain one or more key volumes that are
discriminative and most volumes are irrelevant to the
recognition process. To this end, they proposed a unified deep
learning framework to simultaneously identify discriminative
key volumes and train classifiers, while they discarded all
irrelevant volumes.

The LUPI paradigm was first introduced by Vapnik and Vashist
\cite{Vapnik09} as a new classification setting to model a real
world learning process (i.e., teacher-student learning
relationship) in a max-margin framework, called SVM+. Pechyony
and Vapnik \cite{PechyonyV10} formulated an algorithm for risk
bound minimization with privileged information. Several
variants of the original SVM+ have been proposed in the
literature including SVM+ with $L_1$ regularization
\cite{LNiu12} and multi-task SVM+ \cite{Cai12}. 

Fouad \emph{et al.} \cite{SFouad13} proposed a combination of
privileged information and metric learning. The privileged
information was used to change the metric of the input data and
thus any classifier could be used. Wand and Ji \cite{ZWang15}
also proposed two different loss functions that exploit
privileged information and can be used with any classifier. The
first model encoded privileged information as an additional
feature during training, while the second approach considered
that privileged information can be represented as secondary
labels. 
Wang \emph{et al.} \cite{ZWangGJ14} incorporated privileged
information in a latent max-margin model, where the additional
knowledge was propagated through the latent nodes and the
classification was performed from the regular data. Although
this approach relaxes the strong assumptions of regular and
privileged data relation for classification, it is limited by
the slack variable estimation through SVM optimization. In this
work, we address this problem by replacing the slack variables
for the maximum margin violation and solve the unconstrained
soft-margin SVM optimization problem.

Serra-Toro \emph{et al.} \cite{SerraToro14} proved that
successfully selecting information that can be treated as
privileged is not a straightforward problem. The choice of
different types of privileged information in the context of an
object classification task implemented in a max-margin scheme
was also discussed in \cite{SharmanskaQL13}. Both regular and
privileged features were considered of equivalent difficulty
for recognizing the true class. Wang \emph{et al.}
\cite{ShangfeiWang2015} proposed a Bayesian network to learn
the joint probability distribution of input features, output
target, and privileged information. A combination of the LUPI
framework and active learning has also been explored by Vrigkas
\emph{et al.} \cite{MVrigkas_ICIP16} to model human activities
in a semi-supervised scheme. Recently, the LUPI paradigm has
been employed with applications on gender classification,
facial expression recognition, and age estimation
\cite{kakadiaris2016body, vrigkas2016exploiting,
Wang_2015_16756}.

\section{Methodology}
\label{sec:approach}

\begin{figure}[!t]
\centering
\includegraphics[width=0.8\linewidth]{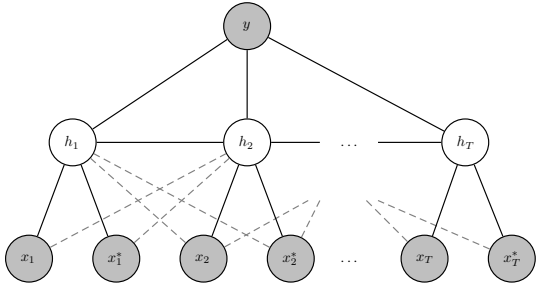}
\caption{Graphical representation of the chain structure model. The
grey nodes are the observed features ($x_{i}$), the privileged
information ($x^{*}_{i}$), and the unknown labels ($y$), respectively.
The white nodes are the unobserved hidden variables ($h$).}
\label{fig:figure2}
\end{figure}

Our method uses HCRFs, which are defined by a chained
structured undirected graph $\mathcal{G} = (\mathcal{V},
\mathcal{E})$ (Fig. \ref{fig:figure2}), as the
probabilistic framework for modeling the behavior of a subject
in a video. During training, a classifier and the mapping from
observations to the label set are learned. In testing, a probe
sequence is classified into its respective state using loopy
belief propagation (LBP) \cite{Komodakis07}.

\subsection{HCRF+ Model Formulation}
\label{subsec:model}

We consider a labeled data set with $N$ video sequences consisting of
triplets $\mathcal{D} =\{(\mathbf{x}_{i,j}, \mathbf{x}^{*}_{i,j},
y_{i})\}_{i=1}^{N}$, where $\mathbf{x}_{i,j} \in
\mathbb{R}^{M_{\mathbf{x}} \times T}$ is an observation sequence of
length $T$ with $j=1 \ldots T$. For example, $\mathbf{x}_{i,j}$ might
correspond to the $j^\text{th}$ frame of the $i^\text{th}$  video
sequence. Furthermore, $y_{i}$ corresponds to a class label defined
in a finite label set $\mathcal{Y}$. In the context of robust
learning using a privileged information paradigm, additional
information about the observations $\mathbf{x}_{i}$ is encoded in a
feature vector $\mathbf{x}^{*}_{i,j} \in
\mathbb{R}^{M_{\mathbf{x}^{*}} \times T}$. Such privileged
information is provided only at the training step and it is not
available during testing. Note that we do not make any assumption
about the form of the privileged data.

In particular, $\mathbf{x}^{*}_{i,j}$ does not necessarily
share the same characteristics with the regular data, but is
rather computed as a very different kind of information, which
may contain verbal and/or non-verbal multimodal cues such as
(i) visual features, (ii) semantic attributes, (iii) textual
descriptions of the observations, (iv) image/video tags, (v)
human poses, and (vi) audio cues. The goal of LUPI is to use
the privileged information $\mathbf{x}^{*}_{i,j}$ as a medium
to construct a better classifier for solving practical problems
than one would learn without it. In what follows, we omit
indices $i$ and $j$ for simplicity.

The HCRF+ model is a member of the exponential family and the
probability of the class label given an observation sequence is given
by:
\begin{equation}
\label{eq:model.1}
\begin{split}
p(y|\mathbf{x},\mathbf{x}^{*};\mathbf{w})
&=\sum_{\mathbf{h}}\exp\left(E(y,\mathbf{h}|\mathbf{x},\mathbf{x}^{*};\mathbf{w}) -
A(\mathbf{w})\right)\, ,
\end{split}
\end{equation}
where $\mathbf{w} = [\boldsymbol\theta, \boldsymbol\omega]$ is
a vector of model parameters, and \mbox{$\mathbf{h} =
\{h_{1},h_{2},\ldots,h_{T}\}$}, with $h_{j} \in \mathcal{H}$
being a set of latent variables. In particular, the number of
latent variables may be different from the number of samples,
as $h_{j}$ may correspond to a substructure in an observation.
Moreover, the features follow the structure of the graph, in
which no feature may depend on more than two hidden states
$h_{j}$ and $h_{k}$ \cite{Quattoni07}. This property not only
captures the synchronization points between the different sets
of information of the same state, but also models the
compatibility between pairs of consecutive states. We assume
that our model follows the first-order Markov chain structure
(i.e., the current state affects the next state). Finally,
$E(y,\mathbf{h}|\mathbf{x};\mathbf{w})$ is a vector of
sufficient statistics and $A(\mathbf{w})$ is the log-partition
function ensuring normalization:
\begin{equation}
\label{eq:model.2}
A(\mathbf{w}) = \log \sum_{y'} \sum_{\mathbf{h}}
\exp\left(E(y',\mathbf{h}|\mathbf{x},\mathbf{x}^{*};\mathbf{w})\right)\, .
\end{equation}

Different sufficient statistics
$E(y|\mathbf{x},\mathbf{x}^{*};\mathbf{w})$ in \eqref{eq:model.1}
define different distributions. In the general case, sufficient
statistics consist of indicator functions for each possible
configuration of unary and pairwise terms:
\begin{equation}
\label{eq:model.3}
\begin{split}
\resizebox{1.0\columnwidth}{!}{$E(y,\mathbf{h}|\mathbf{x},\mathbf{x}^{*};\mathbf{w})
= \displaystyle\sum_{j \in \mathcal{V}}  \Phi(y,h_{j},\mathbf{x}_{j},\mathbf{x}_{j}^{*};\boldsymbol\theta)
 + \displaystyle\sum_{j, k \in \mathcal{E}} \Psi(y,h_{j},h_{k};\boldsymbol\omega)\, ,$}
\end{split}
\end{equation}
where the parameters $\boldsymbol\theta$ and $\boldsymbol\omega$ are
the unary and the pairwise weights, respectively, that need to be
learned. Moreover, the potential functions correspond to the
structure of the graphical model as illustrated in Fig.
\ref{fig:figure2}. For example, a unary potential does not depend on
more than two hidden variables $h_{j}$ and $h_{k}$, and a pairwise
potential may depend on $h_{j}$ and $h_{k}$, which means that there
must be an edge $(j,k)$ in the graphical model.

The unary potential is expressed by:
\begin{equation}
\label{eq:model.3a}
\begin{split}
\Phi(y,h_{j},\mathbf{x}_{j},&\mathbf{x}_{j}^{*};\boldsymbol\theta) =
\sum\limits_{\ell} \phi_{1,\ell}(y,h_{j};\boldsymbol\theta_{1,\ell})\\
& + \phi_{2}(h_{j},\mathbf{x}_{j};\boldsymbol\theta_{2})
+ \phi_{3}(h_{j},\mathbf{x}_{j}^{*};\boldsymbol\theta_{3})\, ,
\raisetag{2.5\baselineskip}
\end{split}
\end{equation}
and it can be seen as a state function, which consists of three
different feature functions. The label feature function, which models
the relationship between the label $y$ and the hidden variables
$h_{j}$, is expressed by:
\begin{equation}
\label{eq:model.3b}
\phi_{1,\ell}(y,h_{j};\boldsymbol\theta_{1,\ell})
= \sum\limits_{\lambda \in \mathcal{Y}} \sum\limits_{a \in \mathcal{H}}
\boldsymbol\theta_{1,\ell} \mathds{1}(y = \lambda) \mathds{1}(h_{j} = a)\, ,
\end{equation}
where $\mathds{1}(\cdot)$ is the indicator function, which is equal
to $1$, if its argument is true and $0$ otherwise. The number of the
label feature functions is $|\mathcal{Y}| \times |\mathcal{H}|$. The
observation feature function, which models the relationship between
the hidden variables $h_{j}$ and the observations $\mathbf{x}_{j}$,
is defined by:
\begin{equation}
\label{eq:model.3c}
\phi_{2}(h_{j},\mathbf{x}_{j};\boldsymbol\theta_{2}) = \sum\limits_{a \in \mathcal{H}}
\boldsymbol\theta_{2}^{\top}\mathds{1}(h_{j} = a) \mathbf{x}_{j}\, .
\end{equation}
The number of the observation feature functions is considered to be
\mbox{$|\mathcal{Y}| \times |M_{\mathbf{x}}|$}. Finally, the
privileged feature function, which models the relationship between
the hidden variables $h_{j}$ and the privileged information
$\mathbf{x}_{j}^{*}$, has $|\mathcal{Y}| \times |M_{\mathbf{x}^{*}}|$
number of functions and is defined by:
\begin{equation}
\label{eq:model.3d}
\phi_{3}(h_{j},\mathbf{x}_{j}^{*};\boldsymbol\theta_{3}) = \sum\limits_{a \in
\mathcal{H}} \boldsymbol\theta_{3}^{\top}\mathds{1}(h_{j} = a) \mathbf{x}_{j}^{*}\, .
\end{equation}

The pairwise potential is a transition function and represents the
association between a pair of connected hidden states $h_{j}$ and
$h_{k}$ and the label $y$. It is expressed by:
\begin{equation}
\label{eq:model.3e}
\Psi(y,h_{j},h_{k};\boldsymbol\omega) = \sum\limits_{\substack{\lambda \in
\mathcal{Y} \\ a,b \in \mathcal{H}}} \sum_{\ell} \boldsymbol\omega_{\ell}
\mathds{1}(y =\lambda) \mathds{1}(h_{j} = a)\mathds{1}(h_{k} = b)\, .
\end{equation}
The number of the transition functions is $|\mathcal{Y}| \times
|\mathcal{H}|^{2}$. HCRF+ keeps a transition
matrix for each label.

\subsection{Maximum Likelihood Learning}
\label{subsec:mlLearning}
In the training step the optimal parameters $\mathbf{w}^{*}$ are
estimated by maximizing the following loss function:
\begin{equation}
\label{eq:model.4}
L(\mathbf{w}) = \sum_{i=1}^{N} \frac{1}{\lambda_{i}}\log p(y_{i}|\mathbf{x}_{i},\mathbf{x}^{*}_{i};\mathbf{w})
- \frac{1}{2\sigma^{2}}\| \mathbf{w} \|^{2}\, .
\end{equation}

The first term is the log-likelihood of the posterior
probability $p(y|\mathbf{x},\mathbf{x}^{*};\mathbf{w})$ and
quantifies how well the distribution in Eq. \eqref{eq:model.1}
defined by the parameter vector $\mathbf{w}$ matches the labels
$y$, while $\lambda$ is a tuning parameter. It can be rewritten
as:
\begin{equation}
\label{eq:model.5}
\begin{split}
\log p(y_{i}|\mathbf{x}_{i},\mathbf{x}^{*}_{i};\mathbf{w}) &= \log \sum_{\mathbf{h}}\exp(
E(y,\mathbf{h}|\mathbf{x}_{i},\mathbf{x}_{i}^{*};\mathbf{w})) \\
&- \log\sum\limits_{\substack{{y'\neq y,\mathbf{h}}}}\exp(E(y',\mathbf{h}|\mathbf{x}_{i},\mathbf{x}_{i}^{*};\mathbf{w}))\, .
\raisetag{2.5\baselineskip}
\end{split}
\end{equation}
The second term  in Eq. \eqref{eq:model.4} is a Gaussian prior
with variance $\sigma^{2}$ and works as a regularizer. The use
of hidden variables makes the optimization of the loss function
non-convex, thus, a global solution is not guaranteed and we
can estimate $\mathbf{w}^{*}$ that are locally optimal. The
loss function  in Eq. \eqref{eq:model.4} is optimized using a
gradient-descent method such as the limited-memory BFGS (LBFGS)
method \cite{Nocedal06}.

\subsection{Maximum Margin Learning}
\label{subsec:mmLearning}
We can easily transform the optimization problem of the loss function
defined in Eq. \eqref{eq:model.4} into a max-margin problem by
substituting the log of the summation over the hidden states and the labels in
Eq. \eqref{eq:model.5} with maximization \cite{Wang11}. The goal is
to maximize the margin between the score of the correct label and the
score of the other labels. To learn the parameters $\mathbf{w}^{*}$
we need to minimize a loss function of the form:
\begin{equation}
\label{eq:model.6}
\begin{split}
L(\mathbf{w}) =  &\sum_{i=1}^{N}\frac{1}{\lambda_{i}}\xi_{i} +
\frac{1}{2\sigma^{2}}\| \mathbf{w} \|^{2} \\
\text{s.t.~}\max\limits_{\substack{{ y'\neq y_{i}},\\\mathbf{h} }} E(y',\mathbf{h}|\mathbf{x}_{i},&\mathbf{x}_{i}^{*};\mathbf{w}) -
\max_{\mathbf{h}} E(y_{i},\mathbf{h}|\mathbf{x}_{i},\mathbf{x}_{i}^{*};\mathbf{w}) \leq \xi_{i} - 1, \\
& \text{and} \,\,\, \xi_{i} \geq 0,  
\raisetag{4.2\baselineskip}
\end{split}
\end{equation}
where parameter $\lambda$ is a tuning parameter. Although we add slack variables $\xi$ to max-margin
optimization, they eventually vanish. We do not estimate the
slacks, but we replace them with the Hinge loss error
\cite{Hastie04} that penalizes the loss when the constraints in
Eq. \eqref{eq:model.6} are violated:
\begin{equation}
\label{eq:model.7}
\begin{split}
\ell_{i}(\mathbf{w}) = \max (0, 1 + (&\max\limits_{\substack{{y'\neq y_{i},\mathbf{h}}}}
E(y',\mathbf{h}|\mathbf{x}_{i},\mathbf{x}_{i}^{*};\mathbf{w}) \\
&- \max_{\mathbf{h}} E(y_{i},\mathbf{h}|\mathbf{x}_{i},\mathbf{x}_{i}^{*};\mathbf{w}) ) ) \, .
\end{split}
\end{equation}

The optimization problem in \eqref{eq:model.6} is equivalent to the
optimization of the following unconstrained problem:
\begin{equation}
\label{eq:model.8}
\begin{split}
L(\mathbf{w}) = \sum_{i=1}^{N}\frac{1}{\lambda_{i}}\ell_{i}(\mathbf{w}) +
\frac{1}{2\sigma^{2}}\| \mathbf{w} \|^{2}\, .
\end{split}
\end{equation}

However, the quantity $\max(0, \cdot)$ is not differentiable and
thus, Eq. \eqref{eq:model.6} is hard to solve. To overcome this
problem we adopt the bundle method \cite{TeoSVL07}, which uses
sub-gradient descent optimization algorithm.

\subsection{Estimation of Regularization Parameters}
\label{subsec:RegularPar}
Both maximum likelihood and max-margin loss functions introduce
regularization parameters that control data fidelity and these
regularization parameters in Eq. \eqref{eq:model.4} and Eq.
\eqref{eq:model.8} may be obtained in closed form. Here, we
examine the case of maximum likelihood optimization as the
estimation of the regularization parameters for the max-margin
optimization is equivalent. We can rewrite the loss function in
Eq. \eqref{eq:model.4} as the sum of individual smoothing
functionals for each of the training samples $N$:
\begin{equation}
\label{eq:model.9}
L(\mathbf{w}) = \sum_{i=1}^{N} \left\{\log p(y_{i}|\mathbf{x}_{i},\mathbf{x}^{*}_{i};\mathbf{w})
- \alpha_{i}(\mathbf{w})\| \mathbf{w} \|^{2} \right\}\, ,
\end{equation}
where $\alpha_{i}(\mathbf{w}) \equiv
\displaystyle\frac{\lambda_{i}}{2\sigma^{2}}$. 

In general, the choice of the regularization parameter for the
optimization of the loss function should be a function of model
parameters $\mathbf{w}$. We consider a linear function $f(\cdot)$
between $\alpha_{i}$ and each term of the loss function:
\begin{equation}
\label{eq:model.11}
\begin{split}
\alpha_{i}(\mathbf{w}) &= f\left(\log p(y_{i}|\mathbf{x}_{i},\mathbf{x}^{*}_{i};\mathbf{w})
- \alpha_{i}(\mathbf{w})\| \mathbf{w} \|^{2} \right) \\
&=\gamma_{i}\left\{\log p(y_{i}|\mathbf{x}_{i},\mathbf{x}^{*}_{i};\mathbf{w})
- \alpha_{i}(\mathbf{w})\| \mathbf{w} \|^{2} \right\} \, ,
\raisetag{2.5\baselineskip}
\end{split}
\end{equation}
where $\gamma$ is determined by the sufficient conditions for
convergence. From Eq. \eqref{eq:model.11}, the regularization parameter $\alpha_{i}$
is computed as:
\begin{equation}
\label{eq:model.11b}
\begin{split}
\alpha_{i}(\mathbf{w}) &= \frac{\log p(y_{i}|\mathbf{x}_{i},\mathbf{x}^{*}_{i};\mathbf{w})}
{\displaystyle\frac{1}{\gamma_{i}} + \| \mathbf{w} \|^{2}} \, ,
\end{split}
\end{equation}
and therefore:
\begin{equation}
\label{eq:model.12}
\frac{1}{\gamma_{i}} > \log p(y_{i}|\mathbf{x}_{i},\mathbf{x}^{*}_{i};\mathbf{w})
- \alpha_{i}(\mathbf{w})\| \mathbf{w} \|^{2} \, .
\end{equation}

We assume that the privileged information is more informative for
classifying human actions than the regular information. Note that,
this is the intuition of using of privileged information as
additional features for classification purposes and it may hold for
most of the cases. Thus, the loss of classifying human actions
directly form $\mathbf{x}$ should be greater or equal than classifying from both
$\mathbf{x}$ and $\mathbf{x}^{*}$:
\begin{equation}
\label{eq:model.13}
\log p(y_{i}|\mathbf{x}_{i};\mathbf{w}) \geq \log p(y_{i}|\mathbf{x}_{i},\mathbf{x}^{*}_{i};\mathbf{w}) \, .
\end{equation}

We can then relax the problem and consider that Eq.
\eqref{eq:model.12} is satisfied when $\tfrac{1}{\gamma_{i}} =\log p(y_{i}|\mathbf{x}_{i};\mathbf{w})$.
Thus, the regularization parameter $\alpha_{i}$ for the loss function
is given by:
\begin{equation}
\label{eq:model.15}
\begin{split}
\alpha_{i}(\mathbf{w}) = \frac{ \log p(y_{i}|\mathbf{x}_{i},\mathbf{x}^{*}_{i};\mathbf{w})}
{\log p(y_{i}|\mathbf{x}_{i};\mathbf{w}) + \| \mathbf{w} \|^{2}} \, ,
\end{split}
\end{equation}
The regularization parameter $\alpha_{i}$ may act as as the
within-classification balance between data and model
parameters. In each step of the optimization process, we
adaptively update the regularization parameter $\alpha_{i}$
providing robustness to the trade-off between the
regularization terms.

Similarly, the regularization parameter $\alpha_{i}$ for the loss
function for the max-margin optimization is given by:
\begin{equation}
\label{eq:model.16}
\alpha_{i}(\mathbf{w}) =  \frac{  \ell_{i}(\mathbf{w})}
{ \zeta_{i}(\mathbf{w}) + \| \mathbf{w} \|^{2}} \, ,
\end{equation}
where $\zeta_{i}(\mathbf{w})$ is the Hinge loss error for
classifying directly from the regular data $\mathbf{x}$:
\begin{equation}
\label{eq:model.17}
\begin{split}
\zeta_{i}(\mathbf{w}) = \max (0, 1 + (&\max\limits_{\substack{{y'\neq y_{i},\mathbf{h}}}}E(y',\mathbf{h}|\mathbf{x}_{i};\mathbf{w}) \\
&- \max_{\mathbf{h}}E(y_{i},\mathbf{h}|\mathbf{x}_{i};\mathbf{w}) ) ) \, .
\raisetag{2.5\baselineskip}
\end{split}
\end{equation}

\subsection{Inference}
\label{subsec:inference}
Having computed the optimal parameters $\mathbf{w}^{*}$ in the
training step, our goal is to estimate the optimal label
configuration over the testing input, where the optimality is
expressed in terms of a cost function. To this end, we maximize the
posterior probability and marginalize over the latent variables
$\mathbf{h}$ and the privileged information $\mathbf{x}^{*}$:
\begin{equation}
\label{eq:model.19}
\begin{split}
\displaystyle y &= \operatorname*{arg\,max}_{y}
p(y|\mathbf{x};\mathbf{w}) \\
&= \operatorname*{arg\,max}_{y}\sum_{\mathbf{h}}\sum_{\mathbf{x}^{*}}
p(y,\mathbf{h},\mathbf{x}^{*}|\mathbf{x};\mathbf{w}) \\
&= \operatorname*{arg\,max}_{y}\sum_{\mathbf{h}}\sum_{\mathbf{x}^{*}}
p(y,\mathbf{h}|\mathbf{x},\mathbf{x}^{*};\mathbf{w})
p(\mathbf{x}^{*}|\mathbf{x};\mathbf{w})\, .
\end{split}
\end{equation}

In the general case, the training samples $\mathbf{x}$ and
$\mathbf{x}^{*}$ may be considered to be jointly Gaussian, thus the
conditional distribution $p(\mathbf{x}^{*}|\mathbf{x};\mathbf{w})$ is
also a Gaussian distribution. In the case of continuous features, the
continuous space of features is quantized to a large number of
discrete values to approximate the true value of the marginalization
of Eq. \eqref{eq:model.19}. However, to efficiently cope with
outlying measurements about the training data, we consider that the
training samples $\mathbf{x}$ and $\mathbf{x}^{*}$ jointly follow a
Student's \textit{t}-distribution. Therefore, the conditional
distribution $p(\mathbf{x}^{*}|\mathbf{x};\mathbf{w})$ is also a
Student's \textit{t}-distribution
$\text{St}(\mathbf{x}^{*}|\mathbf{x};\mu^{*},\Sigma^{*},\nu^{*})$,
where $\mathbf{x}^{*}$ forms the first $M_{{\mathbf{x}}^{*}}$
components of $\left( \mathbf{x}^{*}, \mathbf{x}\right)^{T}$,
$\mathbf{x}$ comprises the remaining $M - M_{\mathbf{x}^{*}}$
components, $\mu^{*}$ is the mean vector, $\Sigma^{*}$ is the
covariance matrix and $\nu^{*} \in [0, \infty)$ corresponds to the
degrees of freedom of the distribution \cite{Kotz04}. Note that by
letting the degrees of freedom $\nu^{*}$ go to infinity, we can
recover the Gaussian distribution with the same parameters. If the
data contain outliers, the degrees of freedom parameter $\nu^{*}$ is
weak and the mean and covariance of the data are appropriately
weighted in order not to take into account the outliers. More details
on how the parameters of the conditional Student's
\textit{t}-distribution $p(\mathbf{x}^{*}|\mathbf{x};\mathbf{w})$ are
estimated can be found in Appendix \ref{AppendixA}.

Although both distributions
$p(y,\mathbf{h}|\mathbf{x},\mathbf{x}^{*};\mathbf{w})$ and
$p(\mathbf{x}^{*}|\mathbf{x};\mathbf{w})$ belong to the exponential
family, the graph in Fig. \ref{fig:figure2} is cyclic, and therefore
an exact solution to Eq. \eqref{eq:model.19} is generally
intractable. For this reason, approximate inference is employed for
estimation of the marginal probability by applying the LBP algorithm
\cite{Komodakis07}.


\section{Experimental Results}
\label{sec:experiments} We evaluated our method on four
challenging publicly available datasets. Three different types
of privileged information were used: audio signal, human pose,
and semantic attribute annotation.

We propose four variants of our approach, called \emph{Maximum
Likelihood LUPI Hidden Conditional Random Field (ml-HCRF+)},
\emph{Adaptive Maximum Likelihood LUPI Hidden Conditional
Random Field (aml-HCRF+)}, \emph{Maximum Margin LUPI Hidden
Conditional Random Field (mm-HCRF+)}, and \emph{Adaptive
Maximum Margin LUPI Hidden Conditional Random Field
(amm-HCRF+)}, depending on which learning method we apply
(i.e., maximum likelihood or max-margin) and whether we
automatically estimate the regularization parameters of the
corresponding loss function or not.

\subsection{Datasets}
\label{subsec:datasets}
\textbf{Parliament} \cite{VrigkasM_SETN14}: This dataset is a
collection of $228$ video sequences, depicting political
speeches in the Greek parliament, at a resolution of $320
\times 240$ pixels at $25$ fps. The video sequences were
manually labeled with one of three behavioral labels:
\emph{friendly}, \emph{aggressive}, or \emph{neutral}.

\textbf{TV human interaction (TVHI)} \cite{Perez12}: This dataset
consists of $300$ video sequences collected from over $20$ different
TV shows. The video clips contain four kinds of interactions: hand
shakes, high fives, hugs, and kisses, equally split into $50$ video
sequences each, while the remaining $100$ video clips do not contain
any of the aforementioned interactions. 

\textbf{SBU Kinect Interaction (SBU)} \cite{Ykiwon12}: This dataset
contains approximately $300$ video sequences depicting
two-person interactions captured by a Microsoft Kinect sensor. The
dataset contains eight different classes including
approaching, departing, pushing, kicking, punching, exchanging
objects, hugging, and shaking hands, which are performed by seven
different persons. It also contains three-dimensional
coordinates of $15$ joints for each person at each frame. 

\textbf{Unstructured social activity attribute (USAA)}
\cite{FuHXG12}: The USAA dataset includes eight different semantic
class videos of social occasions such as birthday party, graduation
party, music performance, non-music performance, parade, wedding
ceremony, wedding dance, and wedding reception. It contains around
$100$ videos per class for training and testing. Each video is
annotated with $69$ attributes, which can be divided into five
broad classes: actions, objects, scenes, sounds, and camera movement.

\subsection{Implementation Details}
\label{subsec:impDet}

\textbf{Feature selection:} For the evaluation of our method,
we used spatio-temporal interest points (STIP) \cite{Laptev05}
as our base video representation. First, we extracted local
space-time features at a rate of $25$ fps using a
$72$-dimensional vector of HoG and $90$-dimensional vector of
HoF feature descriptors \cite{Klaser08} for each STIP, which
captures the human motion between frames. These features were
selected because they can capture salient visual motion
patterns in an efficient and compact way. In addition, for the
TVHI dataset, we also used the provided annotations, which are
related to the locations of the persons in each video clip,
including the head orientations of each subject in the clips,
the pair of subjects who interact with each other, and the
corresponding labels. For our experiments on Parliament and TVHI datasets, we
used audio features as privileged information. More,
specifically, we employed the mel-frequency cepstral
coefficients (MFCC) \cite{Rabiner93} features and their first
and second order derivatives. The audio signal was sampled at
$16$ KHz and processed over $10$ \textit{ms} using a Hamming
window with $25\%$ overlap. The audio feature vector consisted
of a collection of $13$ MFCC coefficients along with the first
and second derivatives forming a $39$ dimensional audio feature
vector.

Furthermore, for the SBU dataset, we used the poses provided by the
dataset as privileged information. In particular, along with the
positions of the locations of the joints for each person in each
frame, we used six more features such as joint distance,
joint motion, plane, normal plane, velocity, and normal velocity as
described by Yun \emph{et al.} \cite{Ykiwon12}. As a basic
representation of the video data, we used the STIP features.

Finally, we used the USAA dataset and the provided attribute
annotation as privileged information to characterize each class not
with an individual label, but with a feature vector of semantic
attributes. As a representation of the video data, we used the
provided low-level features, which correspond to SIFT \cite{Lowe04},
STIP, and MFCC features. Table \ref{Tab:table1} summarizes all forms
of features used either as regular or privileged for each dataset in
our algorithm during training and testing.

\begin{table}[!t]
\renewcommand{\arraystretch}{1.3}
\caption{Types of features used for human activity recognition for
each dataset. The numbers in parentheses indicate the dimension of
the features. The checkmark corresponds to the usage of the specific
information as regular or privileged. Privileged features are used
only during training.} \label{Tab:table1}
\centering
\begin{tabular}{llcc}
\hline
Dataset & Features (dimension) & Regular & Privileged \\
\hline \hline
\multirow{2}{*}{Parliament \cite{VrigkasM_SETN14}} & STIP ($162$) & \cmark & ~ \\
~ & MFCCs ($39$) & ~ & \cmark\\
\hline
\multirow{3}{*}{TVHI \cite{Perez12}} & STIP ($162$) & \cmark & ~  \\
~ & Head orientations (2) & \cmark & ~ \\
~ & MFCC ($39$) & ~ & \cmark\\
\hline
\multirow{2}{*}{SBU \cite{Ykiwon12}} & STIP ($162$) & \cmark & ~ \\
~ & Pose ($15$) & ~ & \cmark \\
\hline
\multirow{4}{*}{USAA \cite{FuHXG12}} & STIP ($162$) & \cmark & ~ \\
~ & SIFT ($128$) & \cmark & ~ \\
~ & MFCC ($39$) & \cmark & ~\\
~ & Semantic attributes ($69$) & ~ & \cmark \\
\hline
\end{tabular}
\end{table}

\textbf{Model selection:} The proposed model 
was trained by varying the number of hidden states from $3$ to
$20$, with a maximum of $400$ iterations for the termination of
the LBFGS optimization method. The $L_{2}$ regularization scale
term $\sigma$ for the non-adaptive methods was set to $10^{k}$,
with \mbox{$k \in \{-3, \dots, 3\}$}. The evaluation of our
method was performed using $5$-fold cross validation to split
the datasets into training and test sets, and the average
results over all the examined configurations are reported.

\subsection{Multimodal Feature Fusion}
\label{subsec:ConvDiscrCont}
One drawback of combining features of different modalities is
the different probablility distribution that each modality may have. Thus,
instead of directly combining multimodal features together one
may employ canonical correlation analysis (CCA)
\cite{Hardoon04} to exploit the correlation between the
different modalities by projecting them onto a common subspace
so that the correlation between the input vectors is maximized
in the projected space. In this paper, we followed a different
approach. Our model is able to learn the relationship between
the input data and the privileged features. 
To this end, we jointly calibrate the different modalities by
learning a multiple output linear regression model
\cite{PalatucciPHM09}. Let $\mathbf{x} \in \mathbb{R}^{M \times
d}$ be the input raw data and $\mathbf{x}^{*} \in \mathbb{R}^{M
\times p}$ be the set of privileged
features. Our goal is to find a set of weights
$\boldsymbol\gamma \in \mathbb{R}^{d \times p}$, which relates
the privileged features to the regular features by minimizing a
distance function across the input samples and their
attributes:
\begin{equation}
\label{eq:fusion.1}
\operatorname*{arg\,min}_{\boldsymbol\gamma}\| \mathbf{x}\boldsymbol\gamma
- \mathbf{x}^{*} \|^{2} + \eta\| \boldsymbol\gamma \|^{2} \, ,
\end{equation}
where $\| \boldsymbol\gamma \|^{2}$ is a regularization term
and $\eta$ controls the degree of the regularization, which was
chosen to give the best solution by using cross validation with
$\eta \in [10^{-4},1]$. Following a constrained least squares
(CLS) optimization problem and minimizing $\| \boldsymbol\gamma
\|^{2}$ subject to $\mathbf{x}\boldsymbol\gamma = \mathbf{x}^{*}$,
Eq. \eqref{eq:fusion.1} has a closed form solution
$\boldsymbol\gamma = \left(\mathbf{x}^{T}\mathbf{x} + \eta
I\right)^{-1} \mathbf{x}^{T}\mathbf{x}^{*}$, where $I$ is the
identity matrix. Note that the minimization of Eq.
\eqref{eq:fusion.1} is fast since it needs to be solved only
once during training. Finally, we obtain the prediction $f$ of
the privileged features by multiplying the regular features
with the learned weights $f = \mathbf{x} \cdot
\boldsymbol\gamma$. The main steps of the proposed method are
summarized in Algorithm \ref{alg:hcrfplusalgo.1}.

\begin{algorithm}[!t]
\caption{\mbox{Robust privileged probabilistic learning}}
\label{alg:hcrfplusalgo.1}

\SetKwInOut{Input}{Input}
\SetKwInOut{Output}{Output}
\SetAlgoLined

\Input{Training sets $\mathcal{X}$, and
$\mathcal{X^{*}}$, training labels $\mathcal{Y}$}

Perform feature extraction from both $\mathcal{X}$ and
$\mathcal{X^{*}}$\\

Employ Eq. \eqref{eq:fusion.1} and project $\mathcal{X}$
and $\mathcal{X^{*}}$ onto a common space\\

Initialize parameters $\mathbf{w}$ randomly\\

\For{$i \in \{1, \ldots, N$\}}{
    /*\textit{Maximum likelihood or max-margin learning}*/
    Estimate the regularization parameter $\alpha_{i}$ using Eqs.
    \eqref{eq:model.15} or \eqref{eq:model.16}  \\
    $\mathbf{w^{*}} \leftarrow $ Train HCRF+ on triplets $(\mathcal{X}_{i},
    \mathcal{X}^{*}_{i}, \mathcal{Y}_{i})$ 
}

\Output{Estimated models' parameters $\mathbf{w^{*}}$}
\end{algorithm}

\subsection{Evaluation of Privileged Information}
\label{subsec:EvalPrivInf}
\begin{figure}[!t]
\centering
\begin{tabular}{cc}
    \includegraphics[width=0.43\columnwidth]{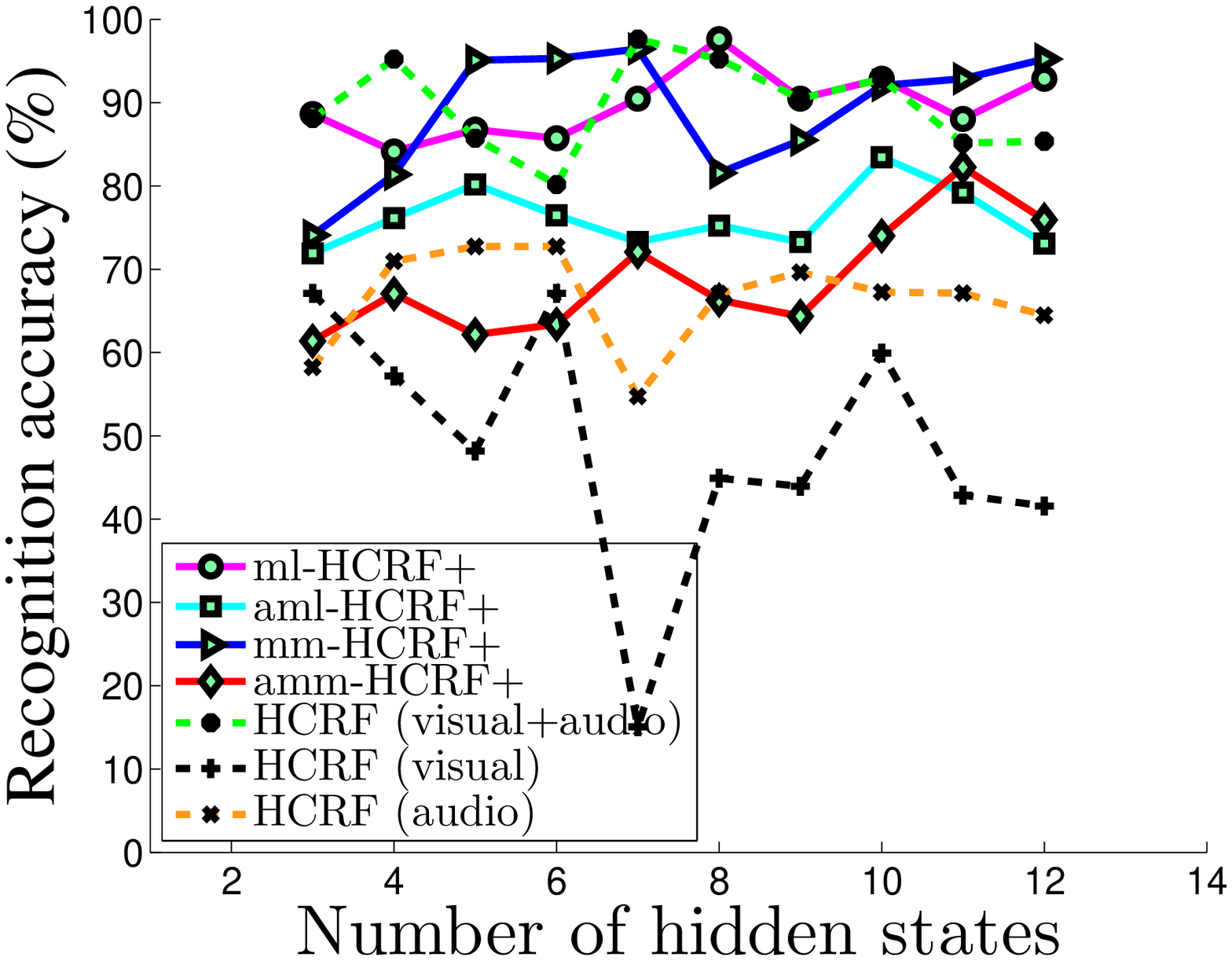} &
    \includegraphics[width=0.43\columnwidth]{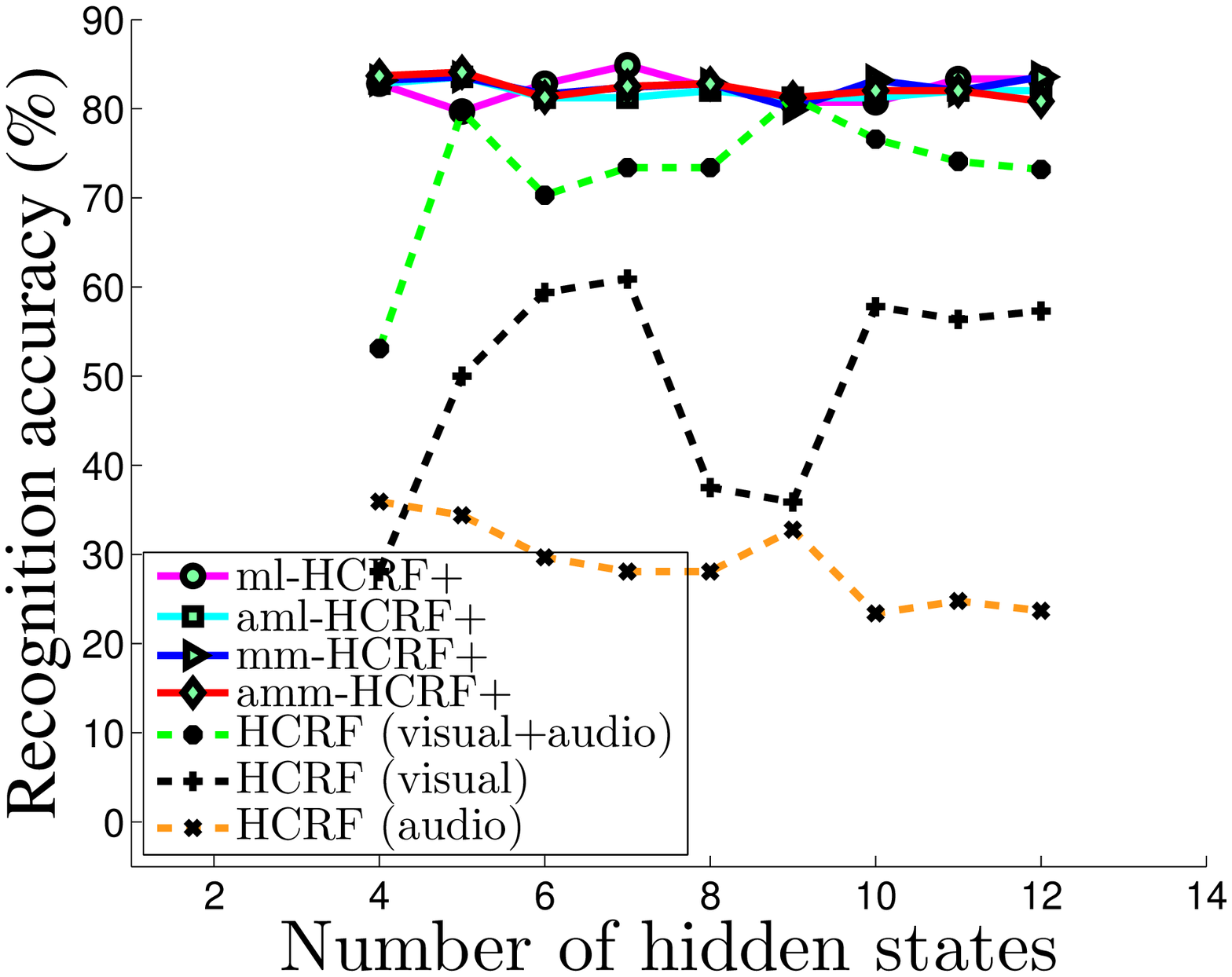} \\
    (a)  & (b)  \\
    \includegraphics[width=0.43\columnwidth]{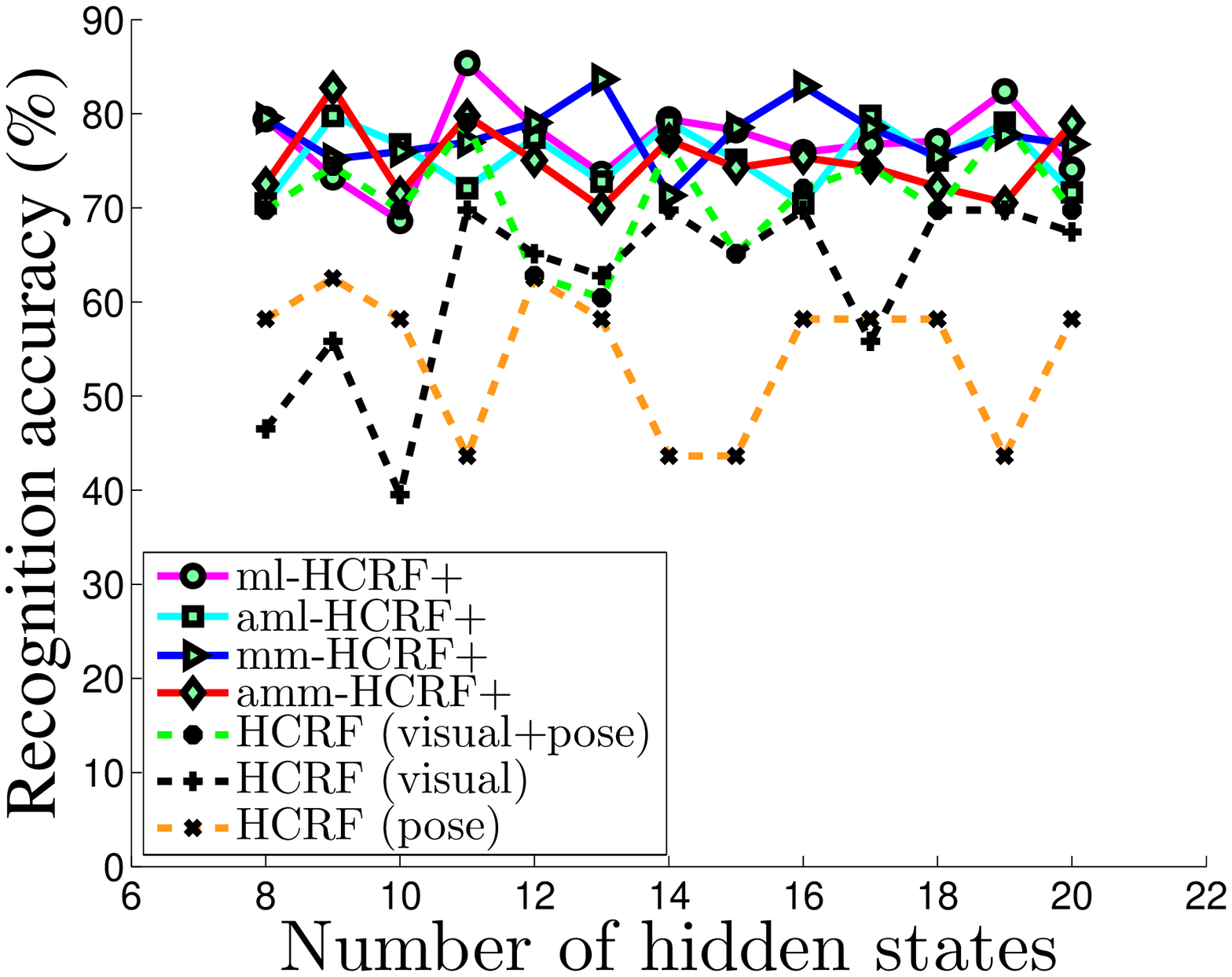} &
    \includegraphics[width=0.43\columnwidth]{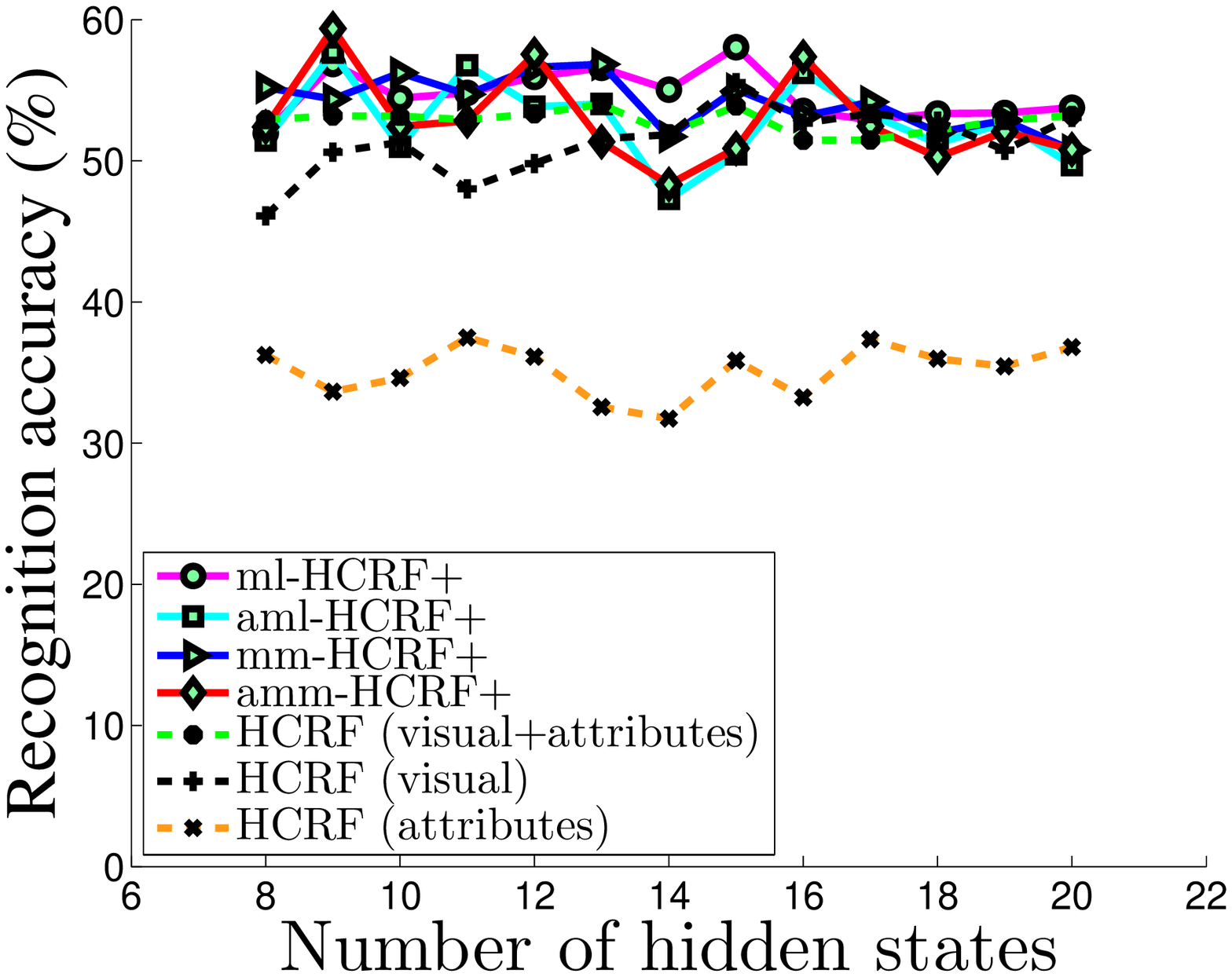} \\
    (c)  & (d)
\end{tabular}
\caption{Comparison of the recognition accuracy of the four different
variants of the proposed method and standard HCRF model with respect
to the number of hidden states for (a) the Parliament
\cite{VrigkasM_SETN14}, (b) the TVHI \cite{Perez12}, (c) the SBU
\cite{Ykiwon12}, and (d) the USAA \cite{FuHXG12} datasets. The text
in parentheses in the legend of each figure corresponds to the type
of information used both for training and testing. }
\label{fig:figure8}
\end{figure}

The classification accuracy with respect to the number of
hidden states is depicted in Fig. \ref{fig:figure8}. We may
observe that all four variants have a similar behavior as the
number of hidden states increases. It is clear that when
privileged information is used, in the vast majority of the
cases ($38$ out of $45$ cases) all variants of HCRF+ perform
better than the standard HCRF model. 
In Fig. \ref{fig:figure8}, the HCRF+ variants and the standard
HCRF model suffer from large fluctuations as the number of
hidden states increases. This is because the number of hidden
states plays a crucial role in the recognition process. Many
hidden states may lead to model overfitting, while few hidden
states may cause underfitting. This would be resolved by the
estimation of the optimal number of hidden states during
learning, but this is not straightforward for this model.
We may also observe that the performance of each modality alone
is kept significantly lower for all configurations of hidden
states, which reinforces the fact that privileged information
may help to construct better classification models.

\begin{figure}[!t]
\begin{center}
\begin{tabular}{cc}
    \includegraphics[width=0.4\columnwidth]{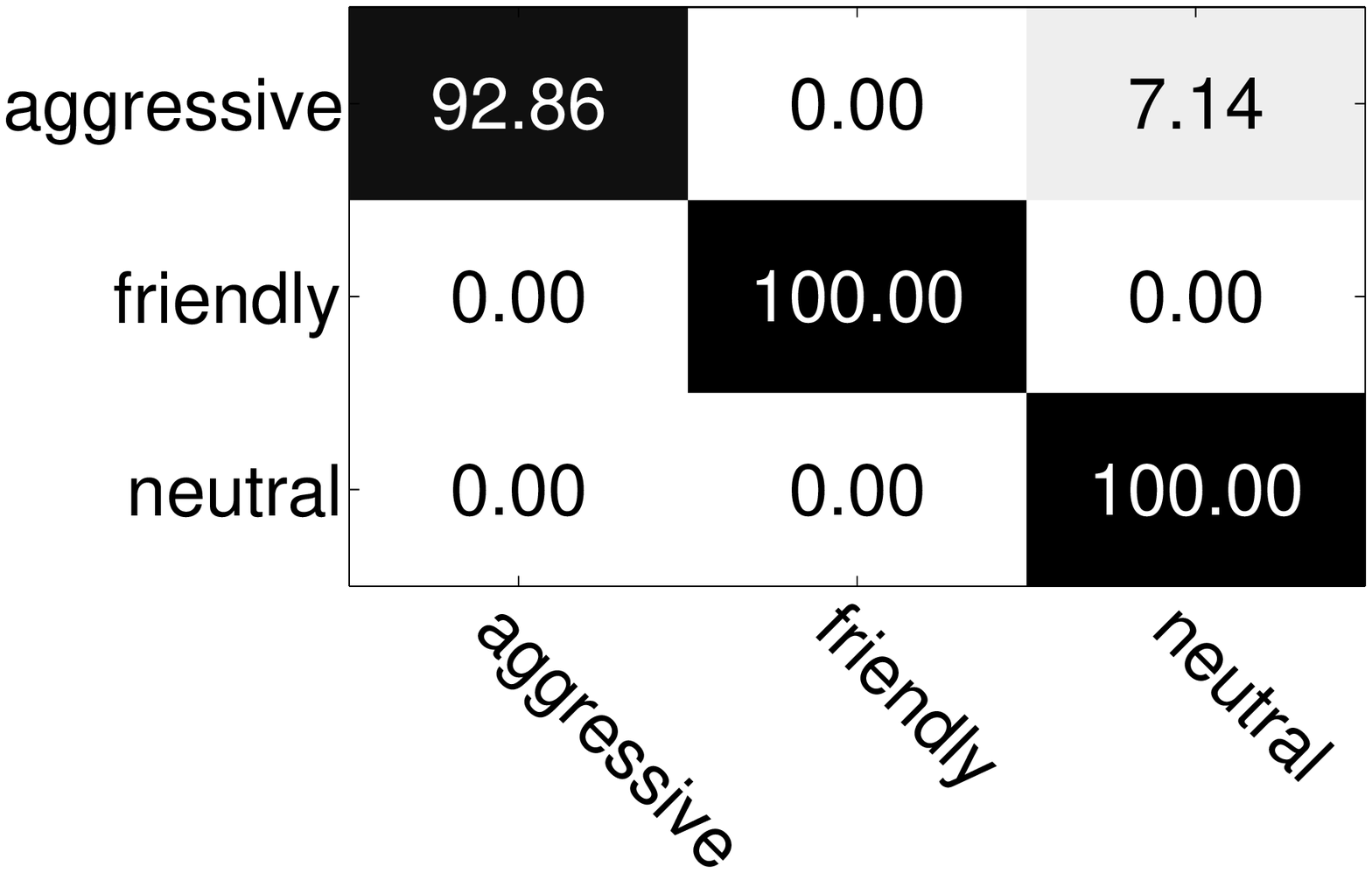} &
    \includegraphics[width=0.4\columnwidth]{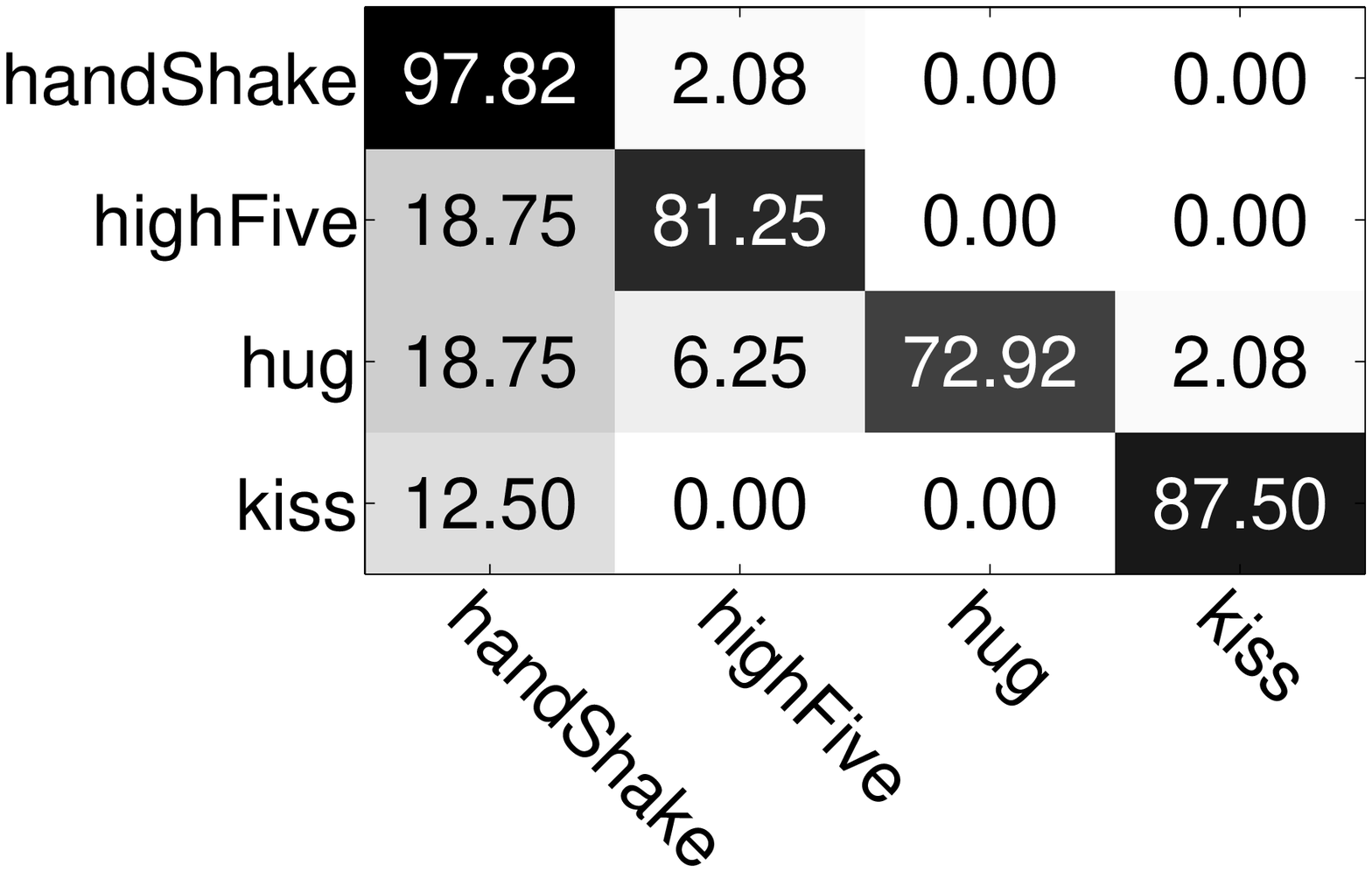} \\
    (a) & (b) \\ 
    \includegraphics[width=0.47\columnwidth]{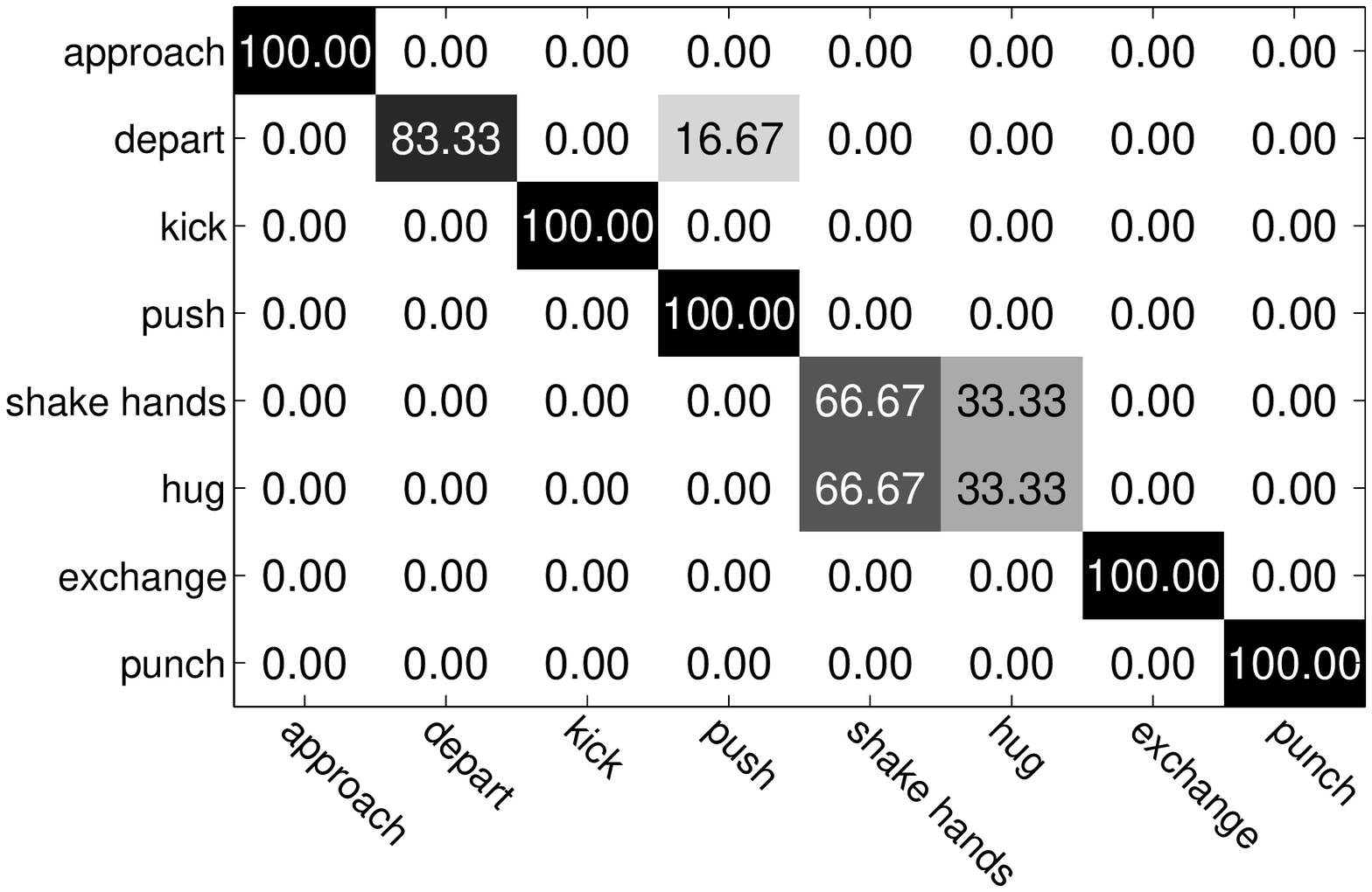} &
    \includegraphics[width=0.48\columnwidth]{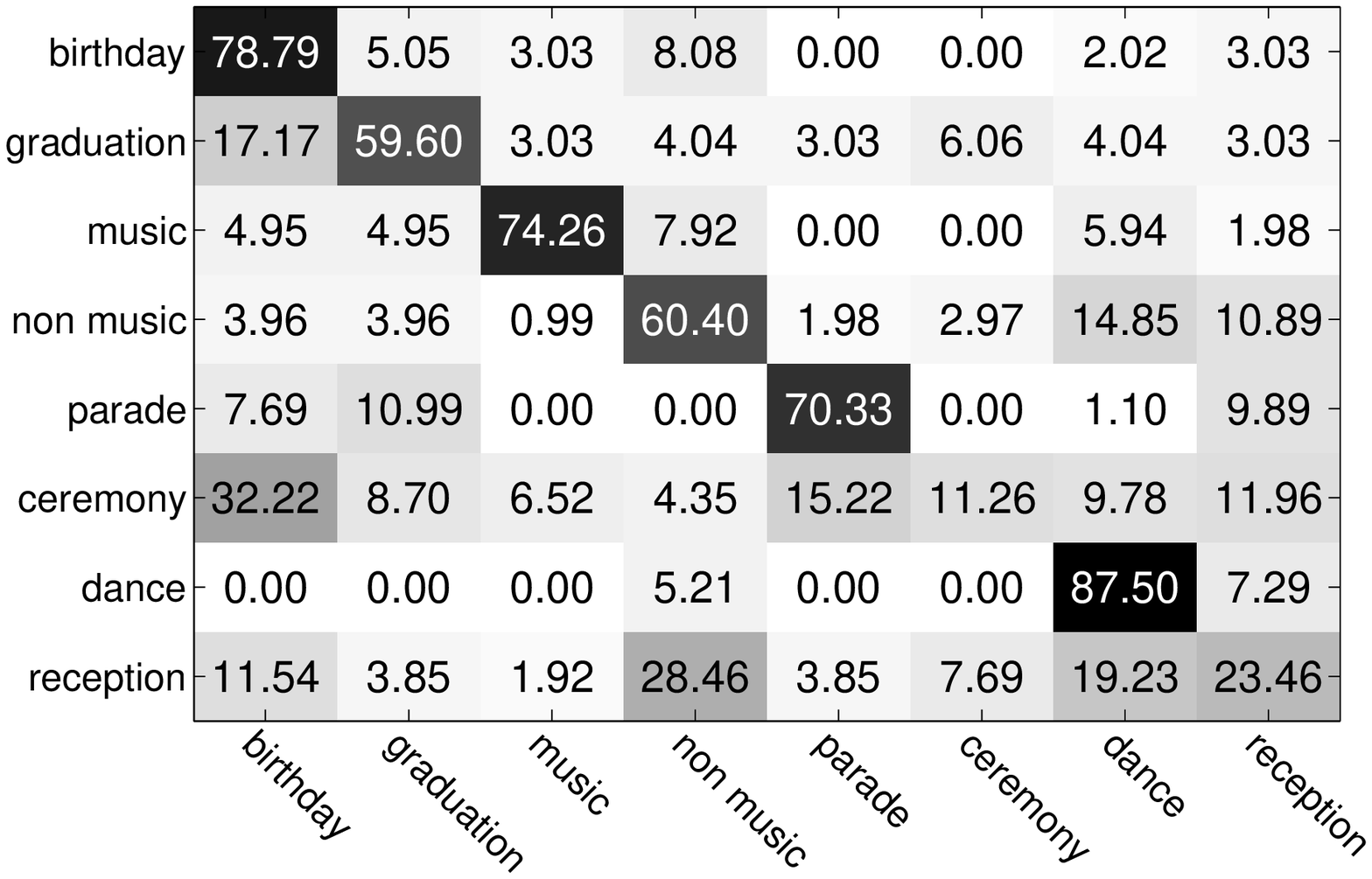} \\
    (c) & (d) 
\end{tabular}
\end{center}
    \caption{Confusion matrices of the
    proposed ml-HCRF+ approach for (a) the Parliament \cite{VrigkasM_SETN14},
    (b) the TVHI \cite{Perez12}, (c) the SBU
    \cite{Ykiwon12}, and (d) the USAA \cite{FuHXG12} datasets.}
\label{fig:figure4}
\end{figure}

The resulting confusion matrices of the best performing variant
are depicted in Fig. \ref{fig:figure4}. It is worth mentioning
that for both the Parliament and the TVHI datasets the
classification errors between different classes
are relatively small.
The SBU dataset has relatively small classification errors, as
only a few classes are confused with each other (e.g., the
class \emph{hugging} versus the class \emph{hand shaking}),
while four out of the eight classes were perfectly recognized.
It is interesting to observe that for the USAA dataset the
different classes may be strongly confused. For example, the
class \emph{wedding ceremony} is confused with the class
\emph{graduation party} and the class \emph{wedding reception}
is confused with the class \emph{non-music performance}. This
is because the different classes may share the same attribute
representation as different videos may have been captured under
similar conditions.

\begin{figure*}[!t]
\centering
\begin{tabular}{cccc}
    \includegraphics[width=0.23\linewidth]{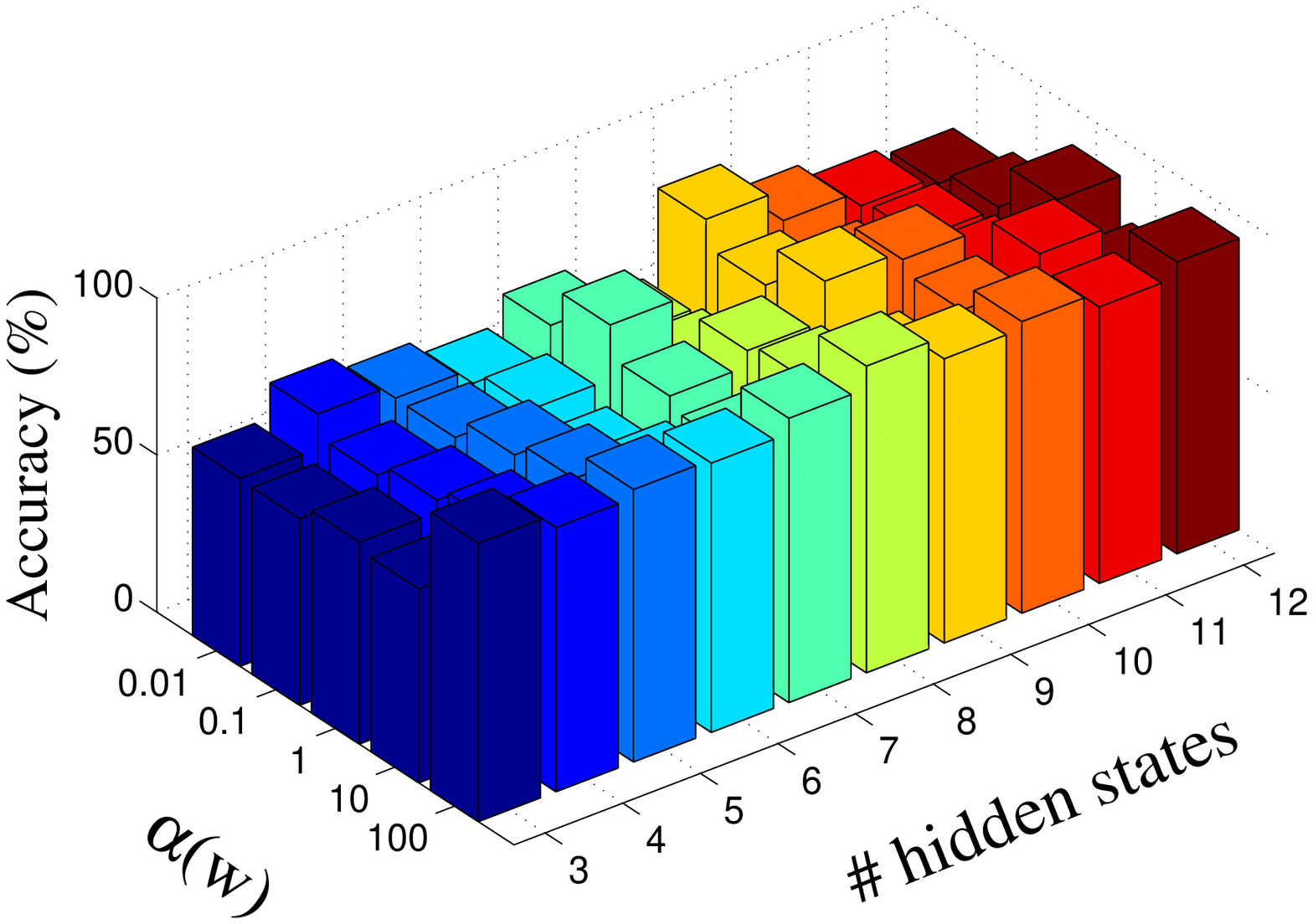}  &
    \includegraphics[width=0.23\linewidth]{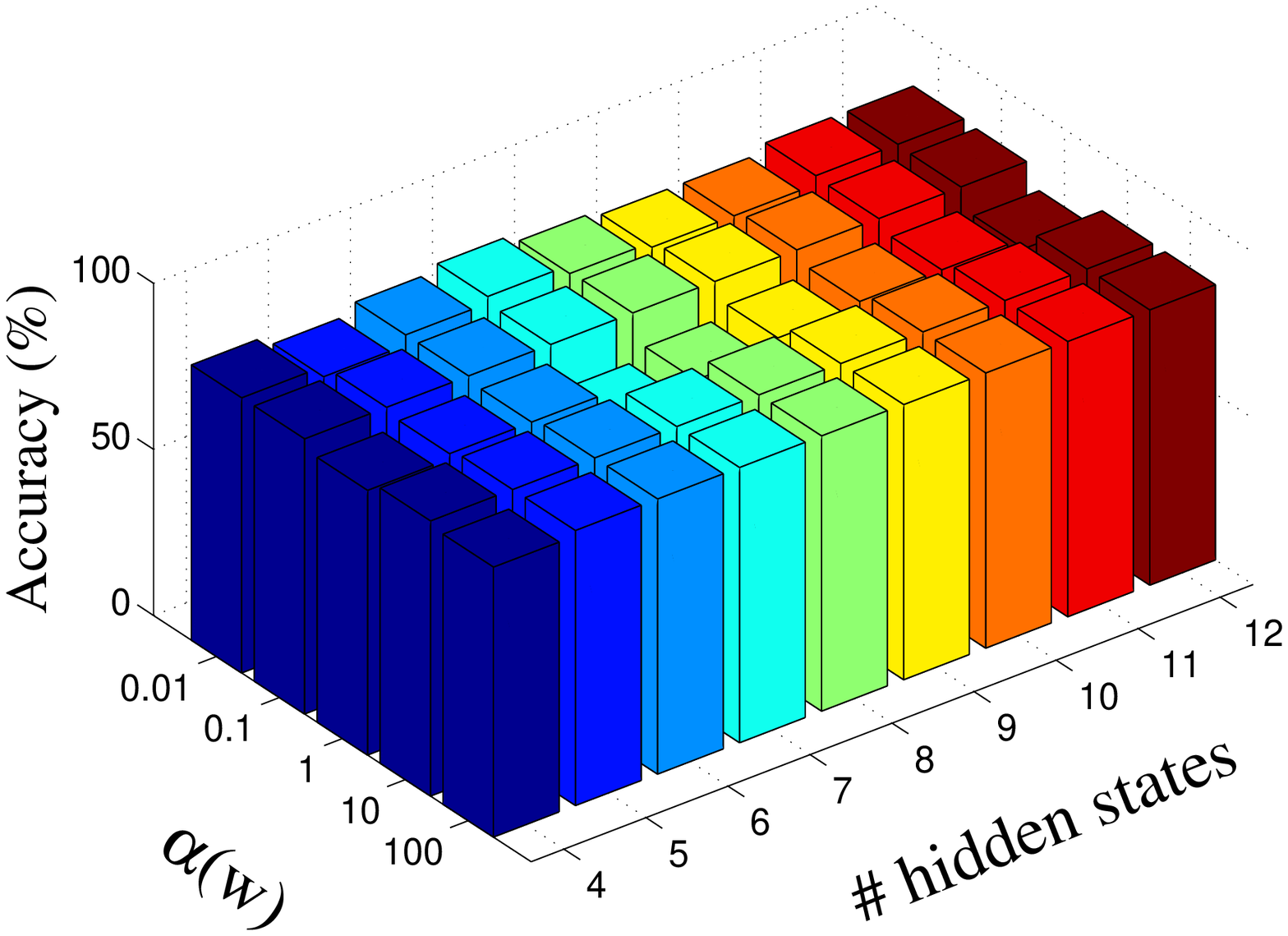} &
    \includegraphics[width=0.23\linewidth]{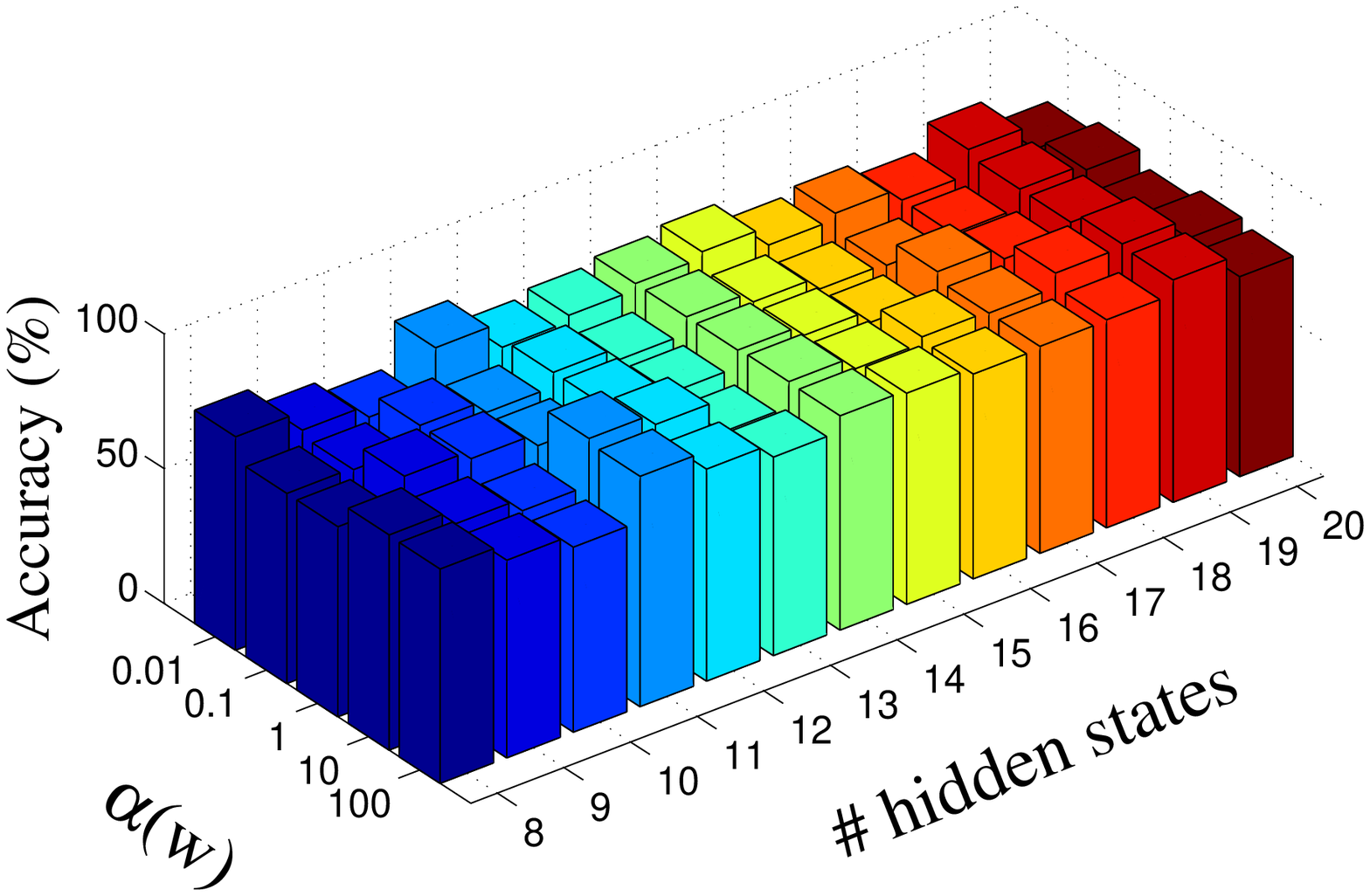}  &
    \includegraphics[width=0.23\linewidth]{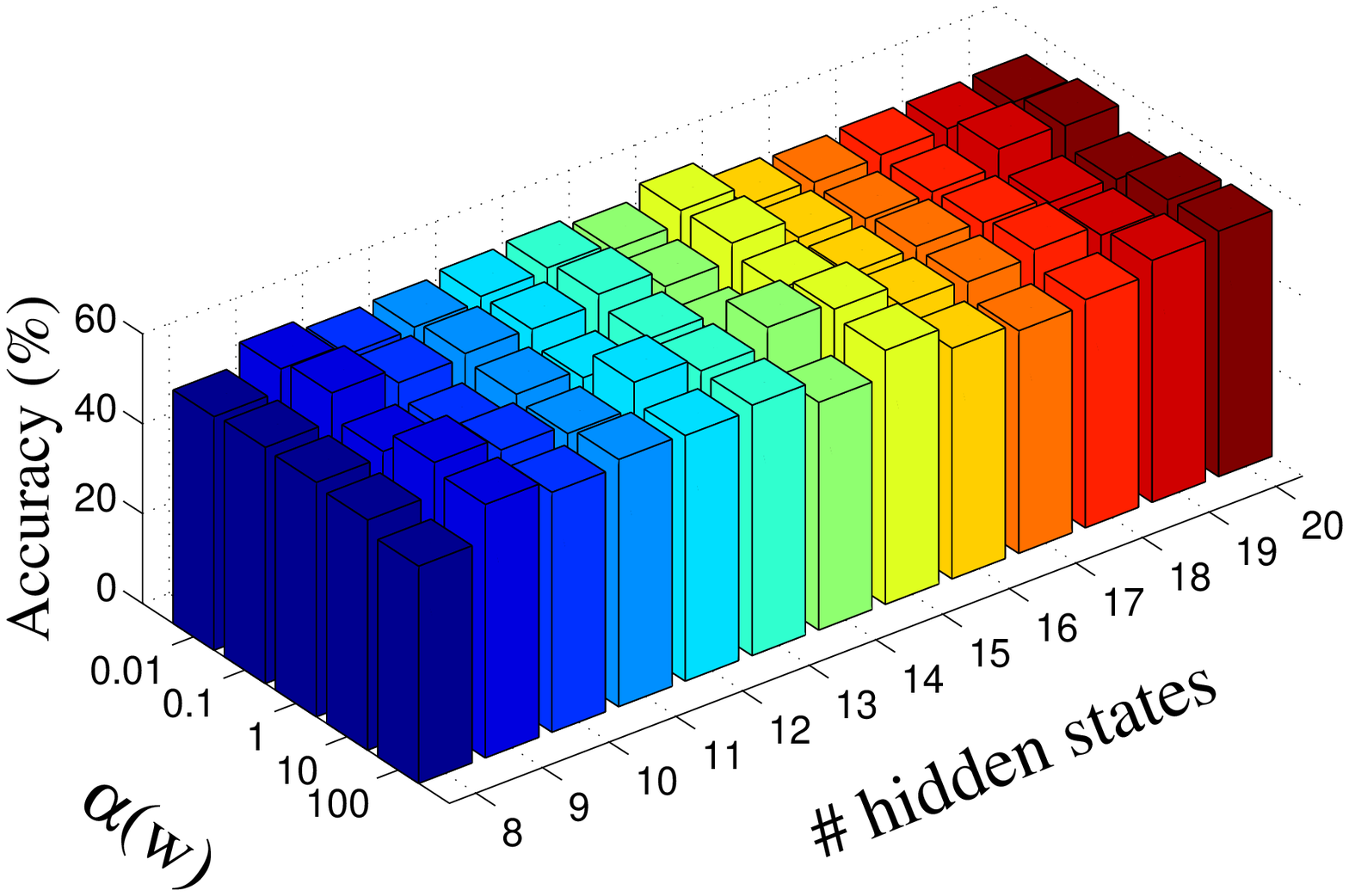}\\
    \includegraphics[width=0.23\linewidth]{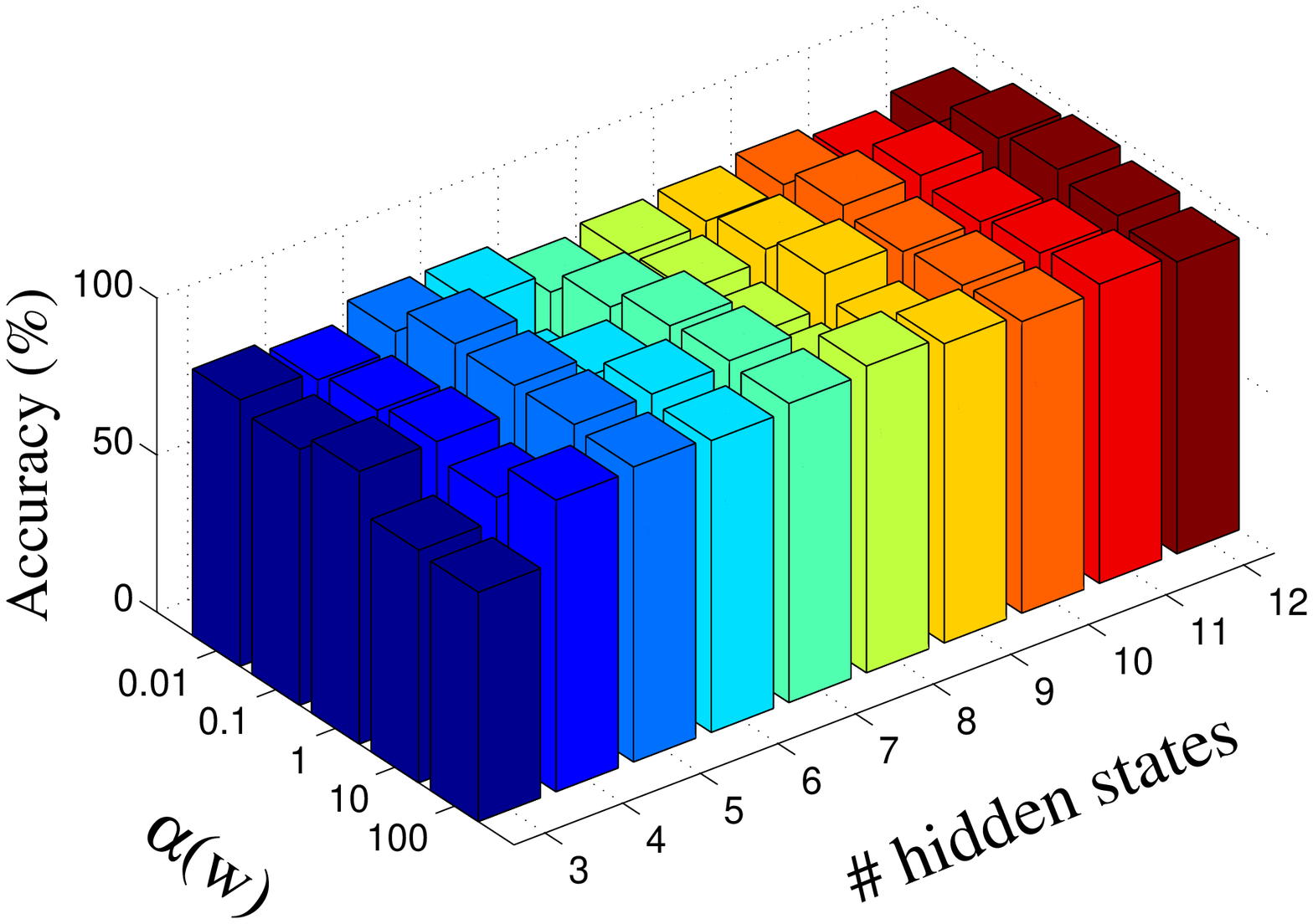}  &
    \includegraphics[width=0.23\linewidth]{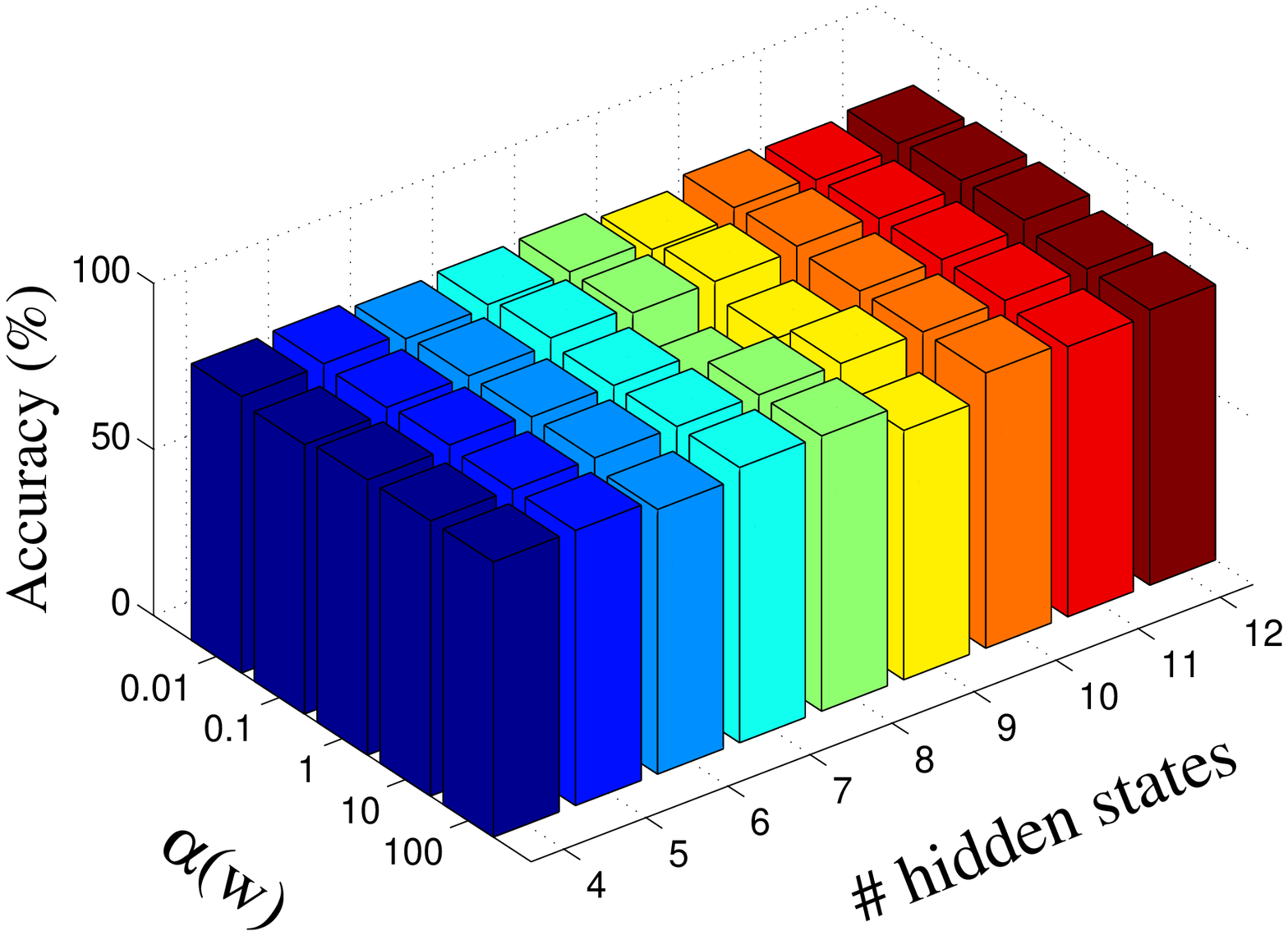} &
    \includegraphics[width=0.23\linewidth]{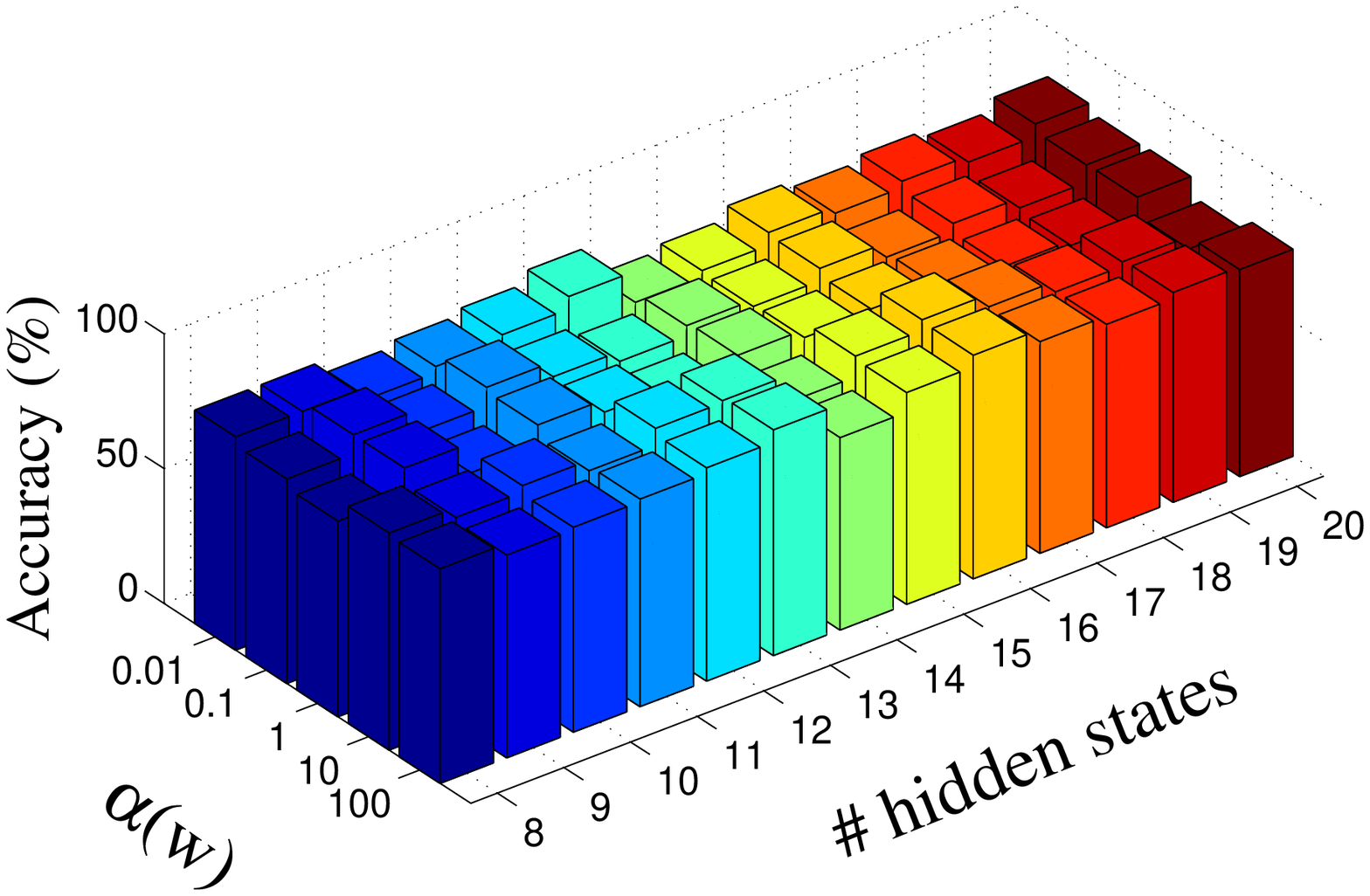}  &
    \includegraphics[width=0.23\linewidth]{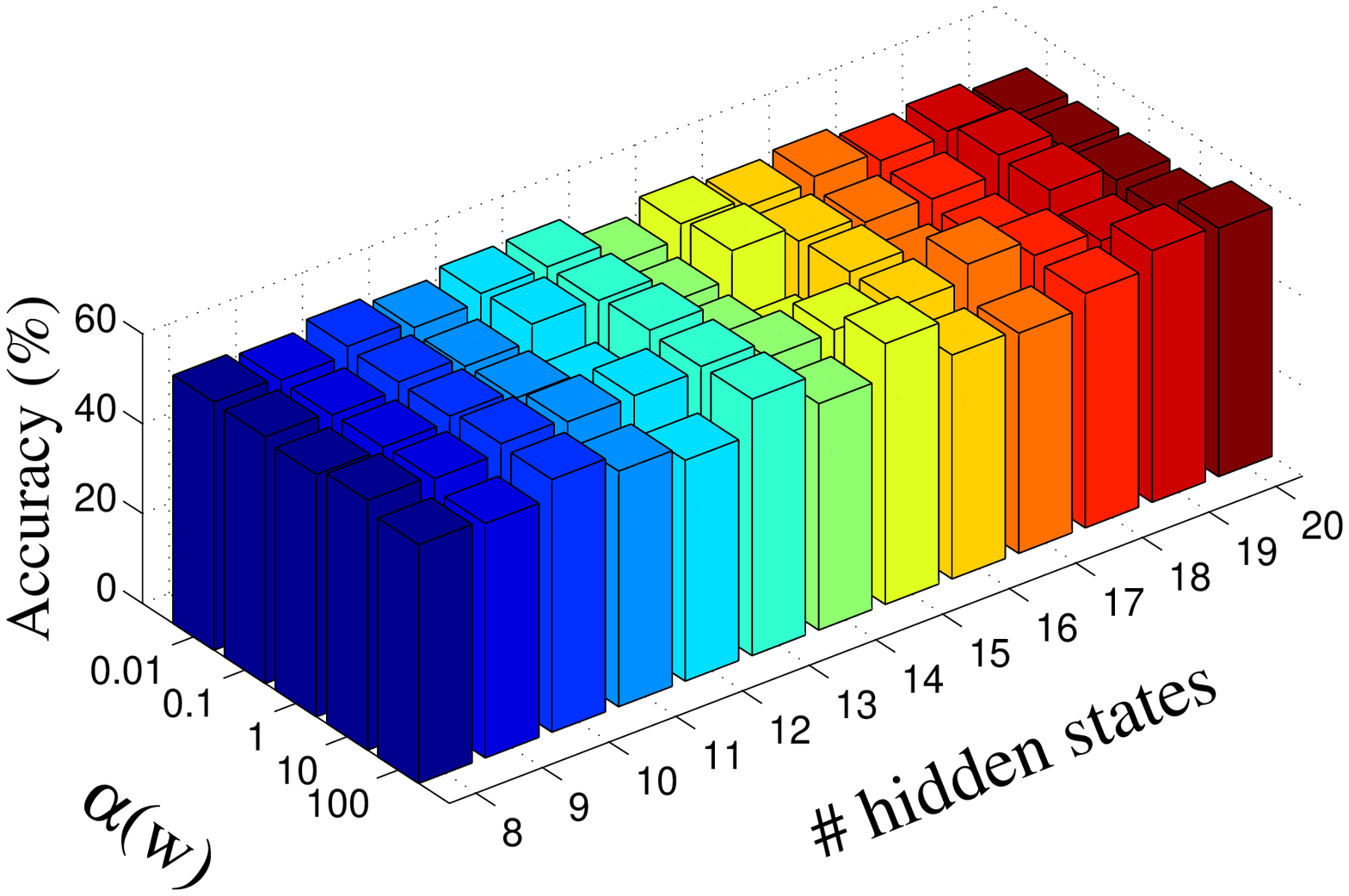}\\
    (a)  & (b) & (c)  & (d)
\end{tabular}
\caption{Recognition performance of the proposed maximum likelihood
(top-row) and max-margin (bottom row) variants as function of the
regularization parameter and the number of hidden states for (a) the
Parliament \cite{VrigkasM_SETN14}, (b) the TVHI \cite{Perez12}, (c)
the SBU \cite{Ykiwon12} and (d) the USAA \cite{FuHXG12} datasets.}
\label{fig:figure6}
\end{figure*}

The behavior of the proposed adaptive model as a function of
the regularization parameters and the number of hidden states
is depicted in Fig. \ref{fig:figure6}. To be consistent to the
non-adaptive methods, the real-valued regularization parameters
were quantized from the continuous to the discrete space with
$\alpha(\mathbf{w}) = 10^{k}, k \in \{-2, \dots, 2\}$ and the
results were averaged. We may observe that the behavior of the
recognition accuracy is smooth for the different values of
$\alpha(\mathbf{w})$ and the number of hidden states, which
indicates that the automatic estimation of $\alpha(\mathbf{w})$
is robust. 

\subsection{Comparisons using Hand-Crafted Features}
\label{subsec:comparison}
In this section, we compare the results of our method with
several state-of-the-art methods. In particular, to show the
benefit of using robust privileged information, we compared our
method both with state-of-the-art methods with and without
incorporating the LUPI paradigm. 
Also, to demonstrate the efficacy of the robust privileged
information to the problem of human activity recognition,
we compared it with ordinary SVM and HCRF, as if they could
access both the regular and the privileged information at test
time. This means that we do not differentiate between regular
and privileged information, but use both forms of information
as regular to infer the underlying class label instead.
Moreover, to complete the study, we also trained an HCRF model
that uses only the regular and only the privileged information
for training and testing. To distinguish between the different
types of information that the HCRF model may use, we
specifically report the type of feature in parentheses after
the HCRF caption. Furthermore, for the SVM+ and SVM we consider
a one-versus-one decomposition of multi-class classification
scheme and average the results for every possible
configuration. Finally, the optimal parameters for the SVM and
SVM+ were selected using cross validation.

\begin{table}[t]
\renewcommand{\arraystretch}{1.3}
\caption{Comparison of the classification accuracies ($\%$) on
the Parliament dataset \cite{VrigkasM_SETN14}. Results
highlighted with \colorbox{blue!15}{light purple} indicate
statistically significant improvement over the second best
method using paired t-test.} 
\label{Tab:table3} \resizebox{\columnwidth}{!}{ \centering
\begin{tabular}{lcccc}
\hline
Method & \multicolumn{1}{c}{Overall} & \multicolumn{1}{c}{Aggressive} & \multicolumn{1}{c}{Friendly} & \multicolumn{1}{c}{Neutral} \\
\hline  \hline
\multicolumn{5}{l}{\emph{Methods without privileged information}} \\
\hline \hline
HCRF (visual+audio) \cite{Quattoni07} & $\mathbf{97.6} \pm 0.6$ & $\,\,\,92.7$ & $100.0$ & $100.0$ \\
HCRF (visual) \cite{Quattoni07} & $67.1 \pm 1.4$ & $\,\,\,50.0$ & $\,\,\,66.7$ & $\,\,\,84.6$ \\
HCRF (audio) \cite{Quattoni07}& $72.7 \pm 1.8$ & $\,\,\,85.7$ & $\,\,\,55.6$ & $\,\,\,76.9$ \\
Wang and Schmid \cite{WangS13a} & $66.6 \pm 0.5$ & $\,\,\,67.9$ & $\,\,\,60.0$ & $\,\,\,71.1$ \\
Vrigkas \emph{et al.} \cite{VrigkasM_SETN14} & $85.5 \pm 1.7$ & $100.0$ & $\,\,\,60.7$ & $\,\,\,95.8$ \\
SVM \cite{Bishop06} & $72.6 \pm 0.4$ & $\,\,\,76.9$ & $\,\,\,69.8$ & $\,\,\,71.1$ \\
\hline \hline
\multicolumn{5}{l}{\emph{Methods with privileged information}} \\
\hline \hline
SVM+ \cite{Vapnik09} & $78.4 \pm 0.2$ & $\,\,\,77.5$ & $\,\,\,68.9$ & $\,\,\,88.7$ \\
Wang and Ji \cite{ZWang15} & $59.2 \pm 0.2$ & $\,\,\,77.9$ &  $\,\,\,39.2$ & $\,\,\,60.5$ \\
Wang \emph{et al.} \cite{ZWangGJ14} & $96.9 \pm 1.1$ & $\,\,\,90.7$ &  $100.0$ & $100.0$ \\
Sharmanska \emph{et al.} \cite{SharmanskaQL13} & $57.7 \pm 0.4$ & $\,\,\,57.1$ & $\,\,\,58.1$  & $\,\,\,57.8$ \\
\textbf{ml-HCRF+} & \cellcolor{blue!15}$\mathbf{97.6} \pm 0.7$ & $\,\,\,92.9$ & $100.0$ & $100.0$ \\
\textbf{aml-HCRF+} & $83.5 \pm 1.3$ & $\,\,\,85.7$ & $\,\,\,80.0$ & $\,\,\,84.6$ \\
\textbf{mm-HCRF+} & $96.5 \pm 0.9$ & $\,\,\,92.6$ & $100.0$ & $\,\,\,97.4$ \\
\textbf{amm-HCRF+} & $82.3 \pm 1.3$ & $\,\,\,85.7$ & $\,\,\,61.1$ & $100.0$ \\
\hline
\end{tabular}
}
\end{table}

A comparison of the proposed approach with state-of-the-art 
methods on the Parliament dataset are depicted in Table \ref{Tab:table3}.
proposed ml-HCRF+ method has highest recognition accuracy
($97.6\%$) among the other variants of the proposed model,
while it achieves the same accuracy with the standard HCRF
model. Although the adaptive HCRF+ approaches may perform worse
than the non-adaptive variants, they can still achieve better
results than the majority of the state-of-the-art methods. One
reason for this, is that the estimation of the regularization
parameters for the adaptive variants depends on the input
features. Features that belong to the background may influence
the estimation of the regularization parameters as they may
serve as background noise.

It is also worth mentioning that our method is able to increase
the recognition accuracy by nearly $38\%$ with respect to the
methods of Wang and Ji \cite{ZWang15} and the method of
Sharmanska \emph{et al.} \cite{SharmanskaQL13}, which also
incorporate the LUPI paradigm. This significantly high increase
in recognition accuracy indicates the strength of the proposed
method. Moreover, the performance of the proposed approach is
higher approximately by $19\%$ than the SVM+ model and $25\%$
than the standard SVM approach. The Parliament dataset contains
large intra-class variabilities. For example, the interaction
between an arm lift and the raise in the voice may not
exclusively be combined together as some features may act as
outliers and affect the classification accuracy.

\begin{table}[t]
\renewcommand{\arraystretch}{1.3}
\caption{Comparison of the classification accuracies ($\%$) on
TVHI the dataset \cite{Perez12}. Results highlighted with
\colorbox{blue!15}{light purple} indicate statistically
significant improvement using paired t-test.} 
\label{Tab:table4} \resizebox{\columnwidth}{!}{ \centering
\begin{tabular}{lccccc}
\hline
Method & Overall & Hand Shake & High Five & Hug & Kiss \\
\hline  \hline
\multicolumn{6}{l}{\emph{Methods without privileged information}} \\
\hline  \hline
HCRF (visual+audio) \cite{Quattoni07} & $81.3 \pm 0.7$ & $87.5$ & $56.3$ & $87.5$ & $93.8$ \\
HCRF (visual) \cite{Quattoni07} & $60.9 \pm 1.3$ & $56.3$ & $25.0$ & $87.5$ & $75.0$\\
HCRF (audio) \cite{Quattoni07} & $35.9 \pm 1.5$ & $12.5$ & $12.5$ & $43.8$ & $75.0$\\
Wang and Schmid \cite{WangS13a} & $76.1 \pm 0.4$ & $76.2$ & $74.6$ & $74.8$ & $74.6$ \\
SVM \cite{Bishop06} & $75.9 \pm 0.6$ & $74.6$ & $76.3$ & $75.8$ & $76.3$ \\
\hline \hline
\multicolumn{6}{l}{\emph{Methods with privileged information}} \\
\hline  \hline
SVM+ \cite{Vapnik09} & $75.0 \pm 0.2$ & $74.6$ & $76.3$ & $72.8$ & $76.2$ \\
Wang and Ji \cite{ZWang15} & $74.8 \pm 0.2$ & $74.6$ & $76.3$ & $72.2$ & $76.3$ \\
Wang \emph{et al.} \cite{ZWangGJ14} & $84.4 \pm 1.1$ & $93.8$ & $81.2$ & $75.1$ & $87.5$ \\
Sharmanska \emph{et al.} \cite{SharmanskaQL13} & $65.2 \pm 0.1$ & $78.3$ & $54.8$ & $74.3$ & $53.5$ \\
\textbf{ml-HCRF+} & \cellcolor{blue!15}$\mathbf{84.9} \pm 0.8$ & $97.2$ & $81.3$ & $72.9$ & $87.5$ \\
\textbf{aml-HCRF+} & $83.6 \pm 1.1$ & $93.8$ & $81.3$ & $71.8$ & $87.5$ \\
\textbf{mm-HCRF+} & $83.6 \pm 0.6$ & $93.8$ & $81.3$ & $72.5$ & $87.5$ \\
\textbf{amm-HCRF+} & $82.9 \pm 0.8$ & $93.8$ & $81.3$ & $68.8$ & $87.5$ \\
\hline
\end{tabular}
}
\end{table}

The classification results on the TVHI dataset are demonstrated
in Table \ref{Tab:table4}. For this dataset, we significantly managed
to increase the classification accuracy by approximately
$10\%$, with respect to the LUPI-based SVM+ and Wang and Ji
\cite{ZWang15} approaches, as our approach achieves very high
recognition accuracy ($84.9\%$). The improvement of our method
compared to the method of Sharmanska \emph{et al.}
\cite{SharmanskaQL13} and the methods that do not
use privileged information was even higher. 

\begin{table*}[t]
\renewcommand{\arraystretch}{1.3}
\caption{Comparison of the classification accuracies ($\%$) on
the SBU dataset \cite{Ykiwon12}. Results highlighted with
\colorbox{blue!15}{light purple} indicate statistically
significant improvement using paired t-test.} \label{Tab:table5}
\centering
\begin{tabular}{lccccccccc}
\hline
Method & Overall & Approach & Depart & Kick & Push & Shake Hands & Hug & Exchange Objects & Punch \\
\hline  \hline
\multicolumn{10}{l}{\emph{Methods without privileged information}} \\
\hline  \hline
HCRF (visual+pose) \cite{Quattoni07} & $81.4 \pm 0.8$ & $100.0$ & $\,\,\,33.3$ & $100.0$ & $\,\,\,66.7$ & $\,\,\,66.7$ & $75.0$ & $100.0$ & $\,\,\,83.3$\\
HCRF (visual) \cite{Quattoni07}  & $69.8 \pm 1.1$ & $100.0$ & $100.0$ & $100.0$ & $\,\,\,66.7$ & $100.0$ & $\,\,\,0.0$ & $100.0$ &
$\,\,\,\,\,\,0.0$\\
HCRF (pose) \cite{Quattoni07} & $62.5 \pm 1.3$ & $100.0$ & $\,\,\,\,\,\,0.0$ & $100.0$ & $100.0$ & $\,\,\,\,\,\,0.0$ & $\,\,\,0.0$ & $100.0$ &
$100.0$ \\
Wang and Schmid \cite{WangS13a} & $79.6 \pm 0.4$ & $\,\,\,76.2$ & $\,\,\,74.6$ & $\,\,\,78.6$ & $\,\,\,78.9$ & $\,\,\,81.4$ & $79.2$ & $\,\,\,84.3$ &
$\,\,\,83.5$\\
SVM \cite{Bishop06} & $79.4 \pm 0.4$ & $\,\,\,74.9$ & $\,\,\,67.2$ & $\,\,\,68.7$ & $\,\,\,76.9$ & $100.0$ & $59.4$ & $\,\,\,89.4$ & $100.0$\\
\hline \hline
\multicolumn{10}{l}{\emph{Methods with privileged information}} \\
\hline  \hline
SVM+ \cite{Vapnik09} & $79.4 \pm 0.3$ & $\,\,\,76.4$ & $\,\,\,72.6$ & $\,\,\,73.2$ & $\,\,\,91.5$ & $\,\,\,70.2$ & $73.2$ & $\,\,\,81.4 $ & $100.0$\\
Wang and Ji \cite{ZWang15}  & $62.4 \pm 0.3$ & $\,\,\,79.5$ & $\,\,\,61.4$ & $\,\,\,59.2$ & $\,\,\,60.0$ & $\,\,\,59.7$ & $60.5$ &
$\,\,\,56.4$ & $\,\,\,62.6$\\
Wang \emph{et al.} \cite{ZWangGJ14}  & $83.7 \pm 1.6$ & $100.0$ & $\,\,\,66.7$ & $\,\,\,75.0$ & $\,\,\,66.7$ & $\,\,\,66.7$ & $75.5$ & $100.0$ & $100.0$\\
Sharmanska \emph{et al.} \cite{SharmanskaQL13}  & $56.3 \pm 0.2$ & $\,\,\,51.6$ & $\,\,\,79.2$ & $\,\,\,40.9$ & $\,\,\,60.0$ &
$\,\,\,74.1$ &  $39.9$ & $\,\,\,43.6$ & $\,\,\,61.2$\\
\textbf{ml-HCRF+} & \cellcolor{blue!15}$\mathbf{85.4} \pm 0.4$ & $100.0$ & $\,\,\,83.3$ & $100.0$ & $100.0$ & $\,\,\,66.7$ & $33.3$ & $100.0$ & $100.0 $\\
\textbf{aml-HCRF+}  & $79.8 \pm 1.3$ & $100.0$ & $100.0$ & $\,\,\,75.0$ & $\,\,\,77.8$ & $100.0$ & $50.0$ & $\,\,\,66.7$ & $\,\,\,66.7$\\
\textbf{mm-HCRF+}  & $83.7 \pm 0.5$ & $100.0$ & $\,\,\,75.0$ & $100.0$ & $100.0$ & $\,\,\,66.7$ & $25.0$ & $100.0$ & $100.0$\\
\textbf{amm-HCRF+}  & $82.8 \pm 1.3$ & $100.0$ & $\,\,\,66.7$ & $\,\,\,83.4$ & $\,\,\,66.7$ & $\,\,\,66.7$ & $75.0$ & $100.0$ & $100.0$\\
\hline
\end{tabular}
\end{table*}
The classification accuracies for the SBU dataset are presented
in Table \ref{Tab:table5}. The ml-HCRF+ approach achieved the
highest accuracy ($85.4\%$), where the improvement over the
standard HCRF model is nearly $4\%$. Comparing our method to
methods that do not use privileged information, we increased
the classification accuracy in all cases. An interesting
observation of the non-privileged  HCRF (visual) and HCRF
(pose) methods arises. Despite the fact that for some classes
these methods were able to perfectly recognize the underlying
activity, they completely failed to recognize some of the
classes as the rate of false positives may reach $100\%$.
Considerably high improvements are also reported when comparing
our methods with state-of-the-art methods that employ
privileged information.

\begin{table*}[t]
\renewcommand{\arraystretch}{1.3}
\caption{Comparison of the classification accuracies ($\%$) on
the USAA dataset \cite{FuHXG12}. Results highlighted with
\colorbox{blue!15}{light purple} indicate statistically
significant improvement using paired t-test.} \label{Tab:table6}
\centering
\begin{tabular}{lccccccccc}
\hline
Method & Overall & Birthday & Graduation & Music & Non-music & Parade & Ceremony & Dance & Reception \\
\hline  \hline
\multicolumn{10}{l}{\emph{Methods without privileged information}} \\
\hline  \hline
HCRF (visual+attributes)\cite{Quattoni07} & $54.0 \pm 0.8$ & $79.8$ & $59.6$ & $48.5$ & $68.3$ & $61.5$ & $\,\,\,4.4$ & $69.8$ & $21.2$\\
HCRF (visual) \cite{Quattoni07}  & $55.5 \pm 0.9$ & $74.8$ & $50.5$ & $76.4$ & $50.5$ & $79.1$ & $\,\,\,4.3$ & $80.2$ & $19.2$\\
HCRF (attributes) \cite{Quattoni07} & $37.4 \pm 1.0$ & $22.2$ & $41.4$ & $63.7$ & $47.5$ & $35.2$ & $14.1$ & $56.3$ & $\,\,\,0.0$\\
Wang and Schmid \cite{WangS13a} & $55.6 \pm 0.1$ & $52.8$ & $55.3$ & $57.1$ & $58.3$ & $60.2$ & $49.7$ & $59.6$ & $40.1$\\
SVM \cite{Bishop06}& $47.4 \pm 0.1$ & $47.5$ & $47.9$ & $49.4$ & $45.7$ & $48.7$ & $38.2$ & $36.5$ & $45.9$\\
\hline \hline
\multicolumn{10}{l}{\emph{Methods with privileged information}} \\
\hline  \hline
SVM+ \cite{Vapnik09} & $48.5 \pm 0.1$ & $52.7$ & $49.9$ & $53.3$ & $50.9$ & $51.6$ & $48.7$ & $41.1$ & $32.5$\\
Wang and Ji \cite{ZWang15} & $48.5 \pm 0.2$ & $32.9$ & $44.6$ & $52.7$ & $48.9$ & $52.0$ & $49.4$ & $54.7$ & $53.0$\\
Wang \emph{et al.} \cite{ZWangGJ14} & $55.3 \pm 0.9$ & $58.6$ & $68.7$ & $58.4$ & $67.3$ & $74.7$ & $17.4$ & $75.0$ & $15.4$\\
Sharmanska \emph{et al.} \cite{SharmanskaQL13} & $56.3 \pm 0.2$ & $56.9$ & $47.8$ & $62.0$ & $62.6$ & $67.1$ & $51.8$ & $57.5$ & $44.4$\\
\textbf{ml-HCRF+} & $58.1 \pm 1.4$ & $78.8$ & $59.6$ & $74.3$ & $60.4$ & $70.3$ & $11.3$ & $87.5$ & $23.5$\\
\textbf{aml-HCRF+} & $57.5 \pm 1.4$ & $78.8$ & $57.6$ & $78.2$ & $70.3$ & $67.0$ & $\,\,\,3.3$ & $78.1$ & $23.1$\\
\textbf{mm-HCRF+} & $56.8 \pm 0.6$ & $79.8$ & $63.6$ & $79.2$ & $59.4$ & $54.9$ & $14.6$ & $85.4$ & $17.5$\\
\textbf{amm-HCRF+} & \cellcolor{blue!15}$\mathbf{59.4} \pm 0.7$ & $78.8$ & $61.6$ & $77.2$ & $69.3$ & $69.2$ & $18.3$ & $79.2$ & $21.2$\\
\hline
\end{tabular}
\end{table*}

The classification results for the USAA dataset are summarized
in Table \ref{Tab:table6}. The combination of both raw data and
attribute representation of human activities significantly
outperformed the SVM+ baseline and the method of Wang and Ji
\cite{ZWang15} by increasing the classification accuracy by
approximately $11\%$ for the amm-HCFR+ model. An improvement of
$3\%$ with respect to the methods of Sharmanska \emph{et al.}
\cite{SharmanskaQL13} and Wang \emph{et al.} \cite{ZWangGJ14}
was also achieved. Furthermore, the adaptive variants of the
proposed method perform better than their non-adaptive
counterparts for this dataset. Automatic estimation of the
regularization parameters provides more flexibility to the
model as it allows the model to adjust its behavior according
to the training data.

\subsection{Comparisons using Deep Learning Features}
\label{subsec:CNNcomparison}
\begin{figure*}[t]
\begin{center}
\begin{tabular}{c}
    \includegraphics[width=0.99\linewidth, clip=true]{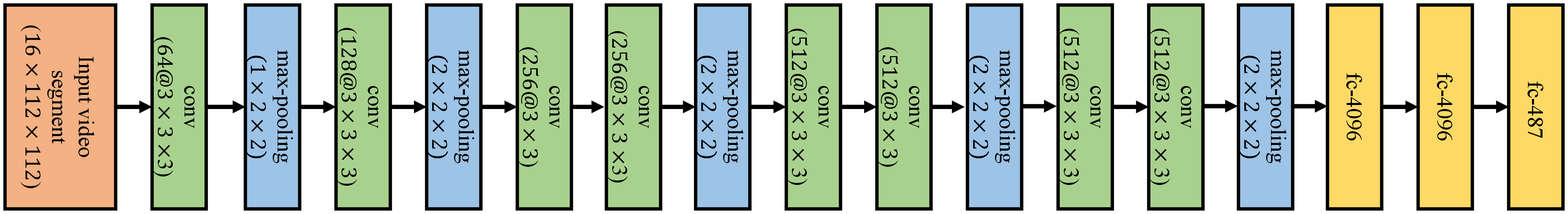} \\
\end{tabular}
\end{center}
\caption{Illustration of the 3D ConvNet architecture
\cite{tran2015learning}. The model has $8$ convolutional
layers, $5$ max-pooling layers and two fully-connected layers
followed by the output layer, which is a softmax classifier
that classifies videos in $487$ categories
\cite{Karpathy_2014_CVPR}. Different colors are used for
different types of layers for better understanding. The net
takes as input a volume of $16$ frames with a height and a
width of size $112$. Convolutional layers perform 3D
convolutions using 3D kernels and max-pooling layers perform 3D
pooling using 3D receptive fields. In convolutional layers, the
number of feature maps is denoted before symbol ``@'' and
follows the size of the kernels. In pooling layers, the size of
each receptive field is shown. In fully-connected layers, it is
denoted the number of hidden units. } \label{fig:figure0}
\end{figure*}

In our experiments, we used CNNs for both end-to-end
classification and feature extraction. We employed the
pre-trained model of Tran \emph{et al.}
\cite{tran2015learning}, which is a 3D ConvNet (Fig.
\ref{fig:figure0}). We selected this model because it was
trained on a very large dataset (Sports 1M
\cite{Karpathy_2014_CVPR}), which provides good features for
the activity recognition task, especially in our case where the
size of the training data is small, making deep learning models
prone to overfitting.

Because both the Parliament and SBU datasets, are fairly small
datasets, only a few parameters had to be trained to avoid
overfitting. Particularly, we replaced the fully-connected
layers of the pre-trained model with a new fully-connected layer
of size $1,024$ and trained the additional layer coupled with a softmax
layer on top of it. For the TVHI dataset, we fine-tuned the
last group of convolutional layers, while for the USAA dataset,
we fine-tuned the last two groups. Each group has two
convolutional layers, while we added a new fully-connected
layer of size $256$ for the TVHI and $1,024$ for the USAA datasets,
respectively. For the optimization process, we used
mini-batch stochastic gradient descent (SGD) with momentum. The
size of the mini-batch was set to $16$ and we used a constant
momentum of $0.9$. For both the Parliament and SBU datasets,
the learning rate was initialized to $0.01$ and it was decayed
by a factor of $0.1$, while the total number of training epochs
was $1,000$. For the TVHI and USAA datasets, we used a constant
learning rate of $10^{-4}$ and the total number of training
epochs was $500$ and $250$, respectively. For all datasets, we
added a dropout layer after the new fully-connected layer with
probability $0.5$. Also, we performed data augmentation on each
batch online and $16$ consecutive frames were randomly selected
for each video. These frames were randomly cropped, resulting
in frames of size $112 \times 112$ and then flipped with
probability $0.5$. For the classification task, we used the
centered $112 \times 112$ crop on the frames of each video
sequence. Then, for each video, we extracted $10$ random clips
of $16$ frames and averaged their predictions. Finally, to
avoid overfitting, we used early stopping and extracted CNN
features from the newly added fully-connected layer.

In addition, we compared the proposed HCRF+ method with the 
LSTM networks \cite{HochreiterS97}, since
it has been proven that they provide good performance in
several sequential classification tasks such as image
description and activity recognition \cite{lrcn2014}. Although
a promising methodology is to train a CNN stacked with an LSTM
layer on top \cite{lrcn2014} for end-to-end feature extraction and sequential classification,
our limited size datasets prevented us on training such a model
due to overfitting. To address this issue, we trained an LSTM
layer with a softmax layer on top, on the features extracted
from the pretrained CNN model. Specifically, we added a dropout layer on
the LSTM's hidden units and an $L_{2}$ regularization on the
softmax units. For estimating the hyperparameters, we performed
a grid search with $5$-fold cross validation to optimize the
learning rate, the number of hidden units, the dropout rate and
the weight decay factor of the $L_{2}$ regularizer. We trained
the LSTM model for $100$ epochs using the Adam optimizer
\cite{KingmaB14} with early stopping.

\begin{table}[t]
\renewcommand{\arraystretch}{1.3}
\caption{Comparison of the classification accuracies ($\%$) on
Parliamment \cite{VrigkasM_SETN14}, TVHI \cite{Perez12}, SBU
\cite{Ykiwon12}, and USAA \cite{FuHXG12} datasets using CNN
features. Results highlighted with \colorbox{blue!15}{light
purple} indicate statistically significant improvement using
paired t-test.} \label{Tab:table7} 
\resizebox{\columnwidth}{!}{ \centering
\begin{tabular}{lcccc}
\hline
Method & Parliament  & TVHI  & SBU  & USAA \\
\hline  \hline
\multicolumn{5}{l}{\emph{Methods without privileged information}} \\
\hline  \hline
HCRF \cite{Quattoni07}  & $84.4 \pm 0.8$ & $89.6 \pm 0.5$ & $91.1 \pm 0.4$ & $91.6 \pm 0.8$ \\
SVM \cite{Bishop06}  & $89.9 \pm 0.5$ & $90.0 \pm 0.3$ & $92.8 \pm 0.2$ & $91.9 \pm 0.3$ \\
CNN (end-to-end) \cite{tran2015learning}  & $78.1 \pm 0.4$ & $60.5 \pm 1.1$ & $94.2 \pm 0.8$ & $67.4 \pm 0.6$ \\
LSTM \cite{HochreiterS97} & $88.3 \pm 0.8$ & $88.4 \pm 1.5$ & $94.7 \pm 0.7$ & $91.3 \pm 1.7$ \\
\hline  \hline
\multicolumn{5}{l}{\emph{Methods with privileged information}} \\
\hline  \hline
SVM+ \cite{Vapnik09}  & $90.0 \pm 0.3$ & $92.5 \pm 0.4$ & $94.8 \pm 0.3$ & $92.3 \pm 0.3$ \\
Wang and Ji \cite{ZWang15} & $83.5 \pm 0.4$ & $88.8 \pm 0.2$ & $92.7 \pm 0.4$ & $92.8 \pm 0.2$ \\
Wang \emph{et al.} \cite{ZWangGJ14} & $84.4 \pm 0.6$ & $85.0 \pm 1.2$ & $91.1 \pm 1.3$ & $93.2 \pm 1.2$ \\
Sharmanska \emph{et al.} \cite{SharmanskaQL13} & $81.8 \pm 0.2$ & $90.0 \pm 0.1$ & $92.9 \pm 0.4$ & $93.5 \pm 0.2$ \\
\textbf{ml-HCRF+} & \cellcolor{blue!15}$\textbf{93.3} \pm 0.7$ & \cellcolor{blue!15}$\textbf{93.2} \pm 0.6$ & \cellcolor{blue!15}$\textbf{94.9} \pm 0.7$ & $93.9 \pm 0.9$ \\
\textbf{aml-HCRF+} & \cellcolor{blue!15}$\textbf{93.3} \pm 0.4$ & $92.5 \pm 1.1$ & $92.9 \pm 0.4$ & $95.9 \pm 1.3$ \\
\textbf{mm-HCRF+} & $88.9 \pm 0.9$ & $92.5 \pm 0.7$ & $93.6 \pm 1.1$ & $95.2 \pm 1.0$ \\
\textbf{amm-HCRF+} & $86.7 \pm 1.2$ & $90.0 \pm 0.8$ & $94.6 \pm 1.0$ & \cellcolor{blue!15}$\textbf{96.4} \pm 1.4$ \\
\hline
\end{tabular}
}
\end{table}

The comparison of the proposed approach with state-of-the-art 
methods using the CNN features is summarized in 
Table \ref{Tab:table7}. The improvement of accuracy with respect to the
hand-crafted feature classification for all datasets (Tables
\ref{Tab:table3} to \ref{Tab:table6}), indicates that CNNs may
efficiently extract informative features without any need to
hand-design them. It is also worth noting that privileged
information works in favor of the classification task in all
cases. The ml-HCRF+ variant achieves the highest results among
all other methods for the Parliament, TVHI, and SBU datasets,
while for the USAA dataset, the amm-HCRF+ variant achieves the
highest recognition accuracy ($96.4\%$). Moreover, the
improvement in accuracy of the proposed model with respect to
the end-to-end CNN classification for the Parliament, TVHI, and
USAA datasets, was approximately $15\%$, $33\%$, and $29\%$,
respectively. This improvement can be explained by the
fact that the CNN model uses a linear classifier in the softmax
layer, while the proposed approach is a more sophisticated
model that can efficiently handle sequential data in a more
principled way. Also, the improvement in performance, brought
by the LSTM compared to the end-to-end classification with the
CNN, validates the ability of LSTMs to capture long-term
dependencies in human activities as LSTMs have memory of
previous activity states and can better model their complex
dynamics. Nonetheless, the proposed model outperforms the LSTM,
for all datasets, a fact that supports our main hypothesis that
the LUPI paradigm may be beneficial for human activity
recognition.

\begin{figure}[t]
\begin{center}
\begin{tabular}{cc}
    \includegraphics[width=0.4\columnwidth]{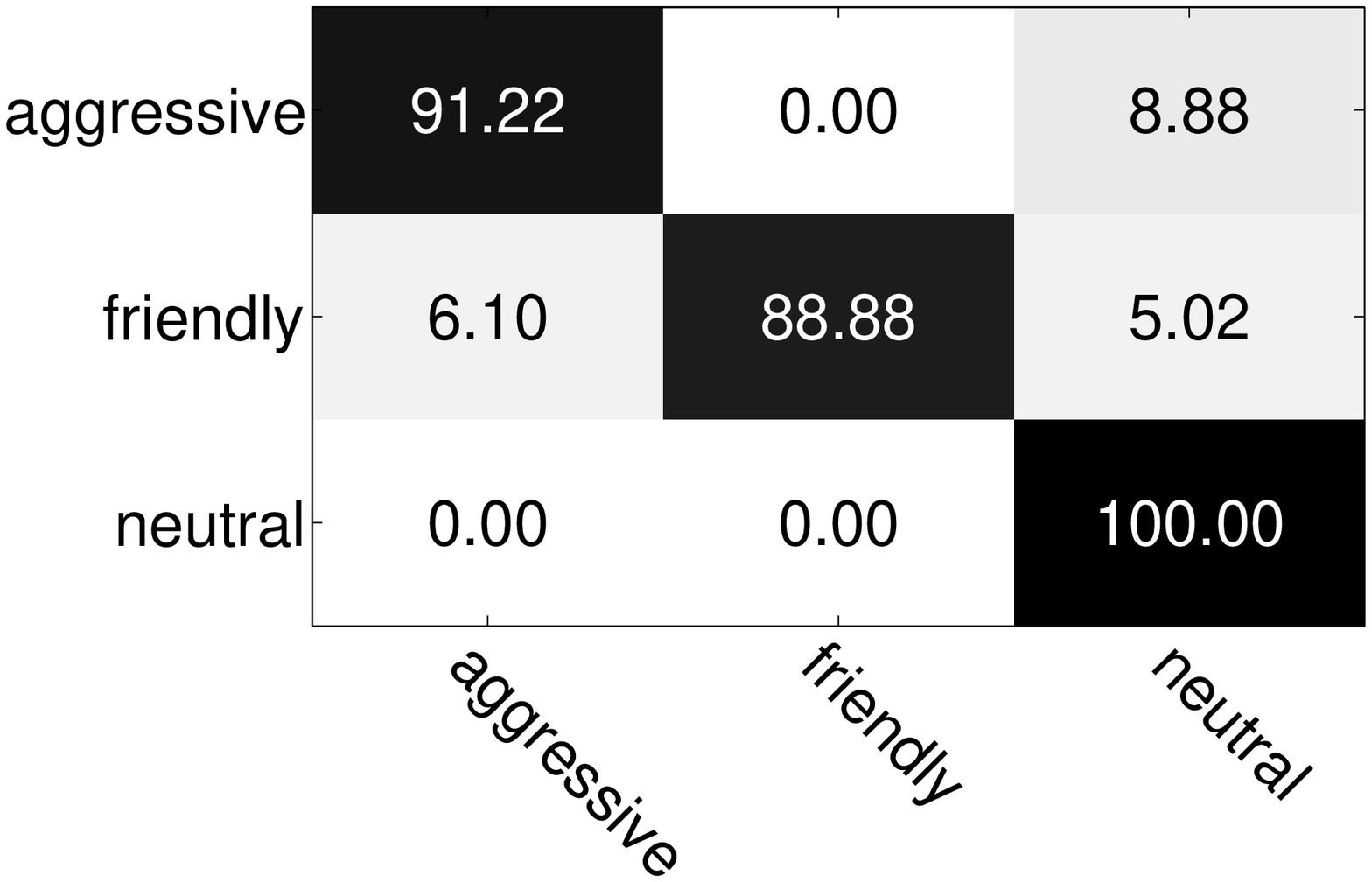} &
    \includegraphics[width=0.4\columnwidth]{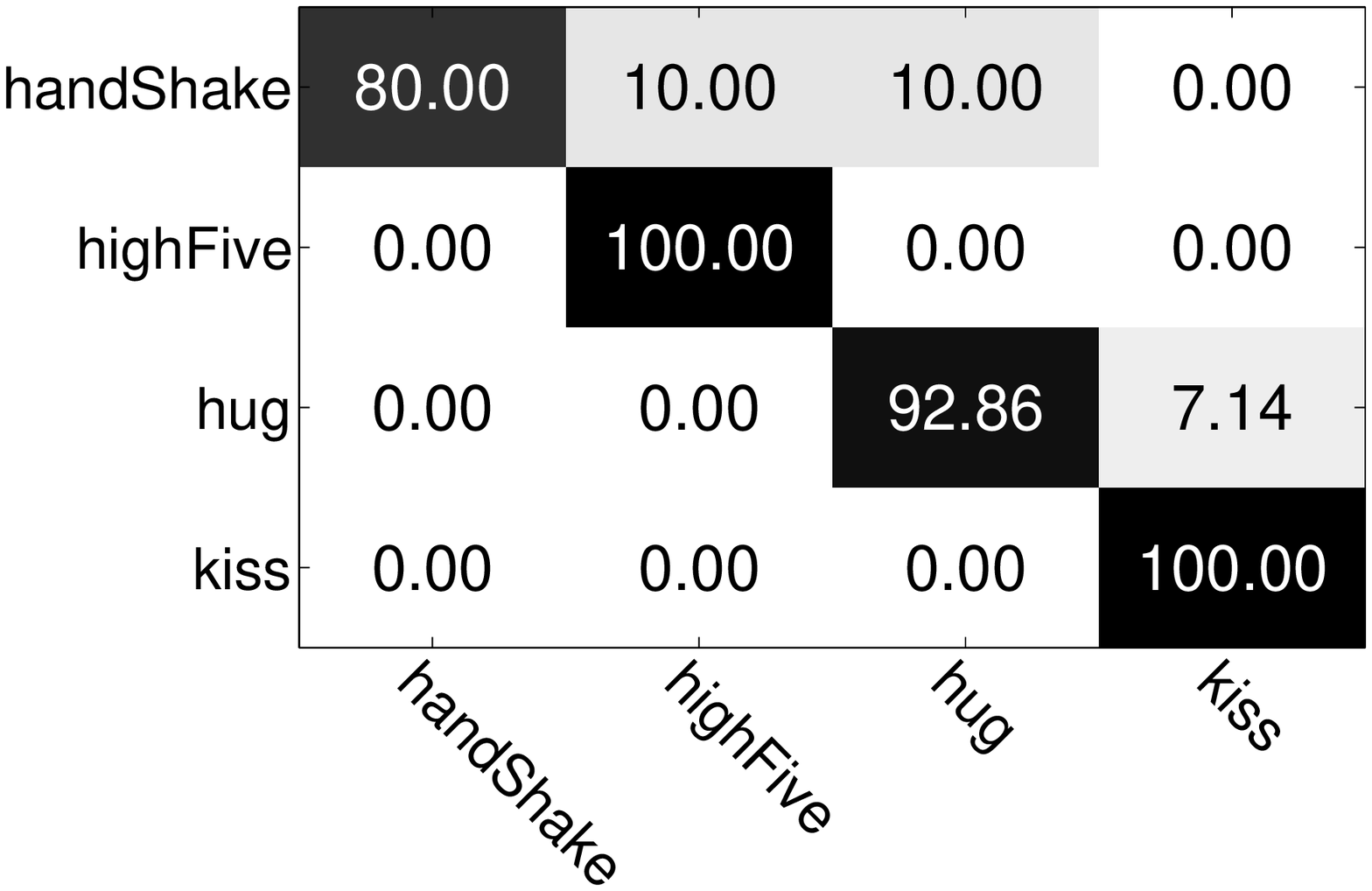} \\
    (a) & (b) \\ 
    \includegraphics[width=0.47\columnwidth]{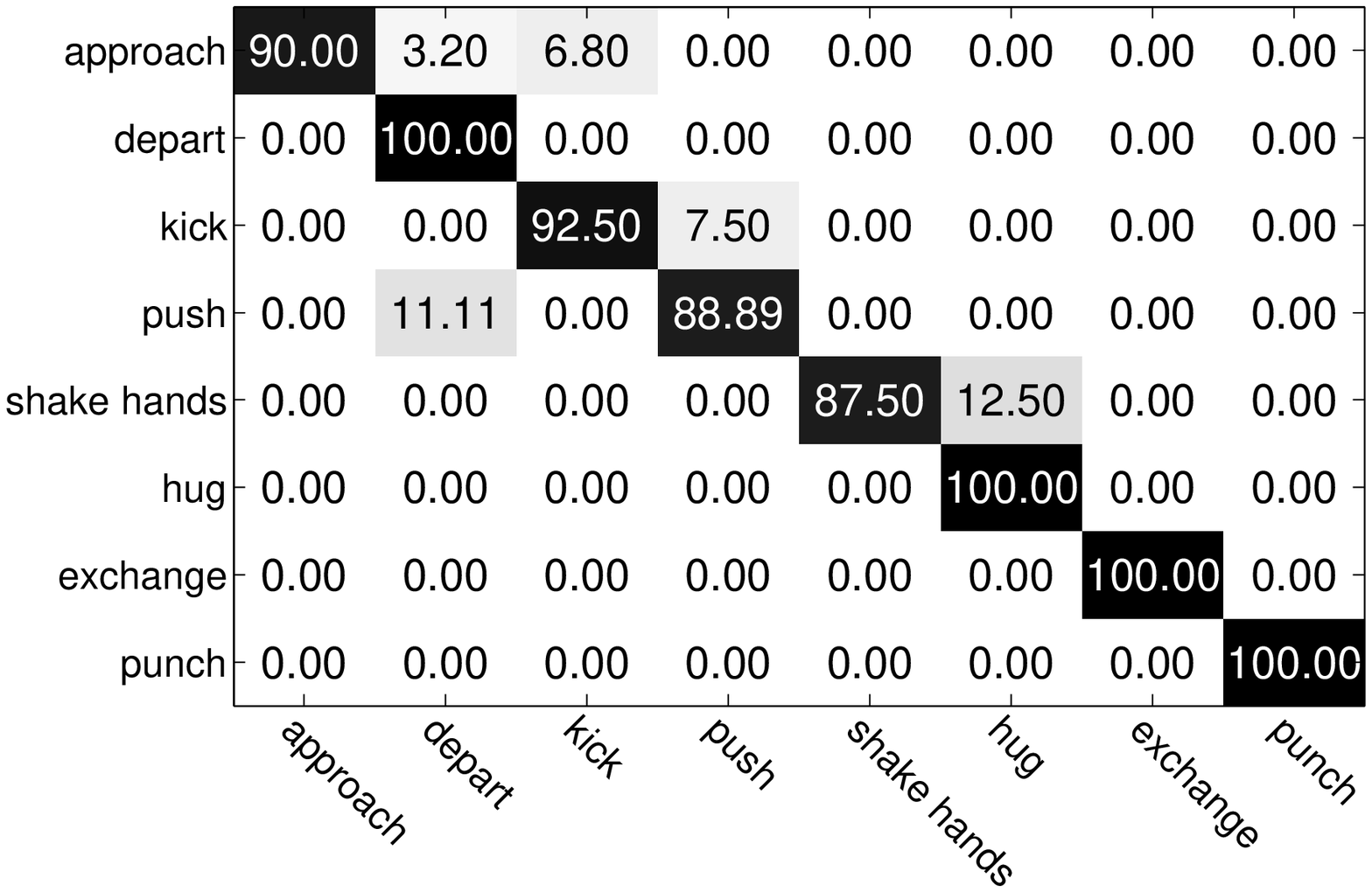} &
    \includegraphics[width=0.48\columnwidth]{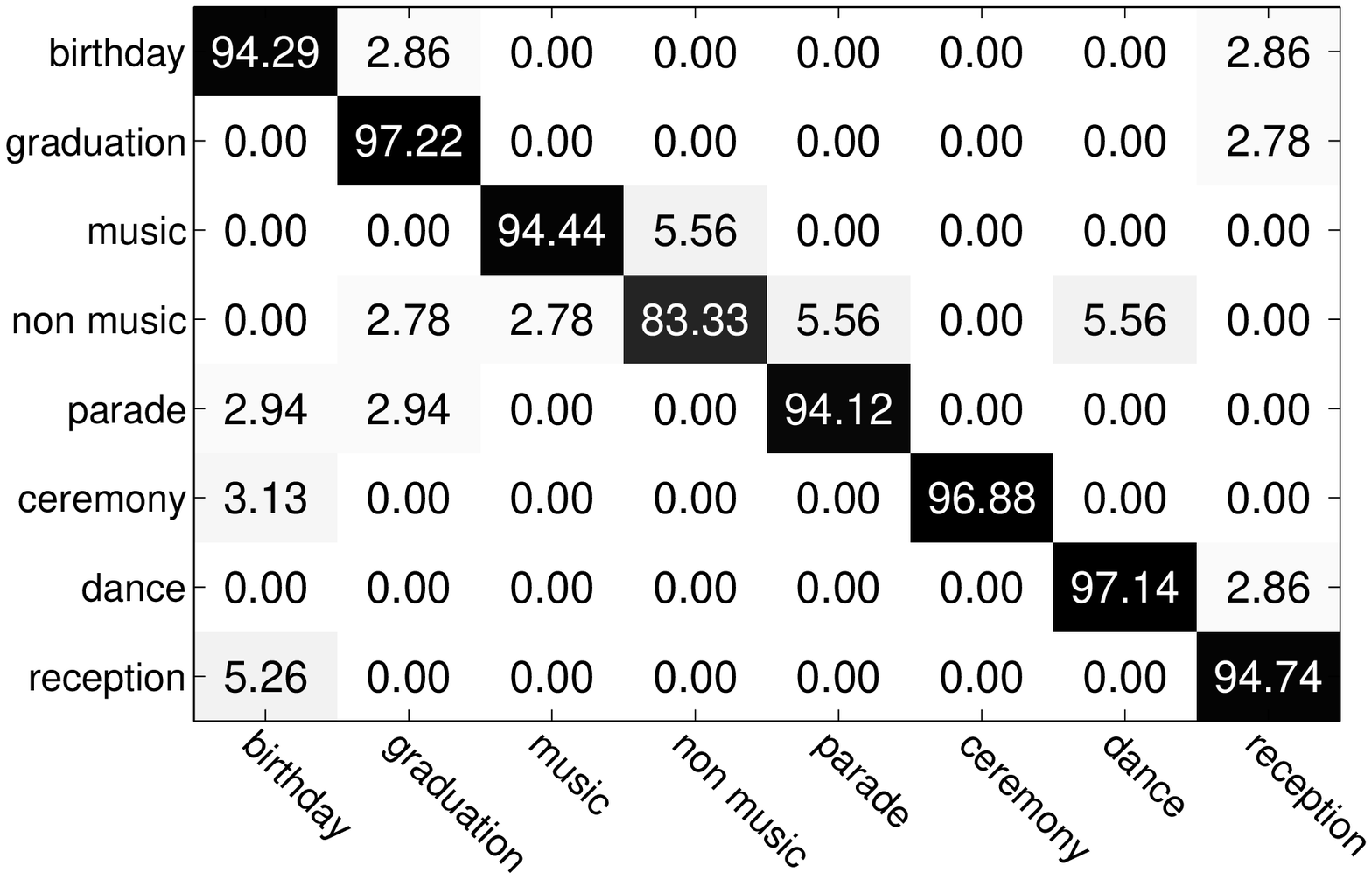} \\
    (c) & (d)                                                           
\end{tabular}
\end{center}
    \caption{Confusion matrices of the
    proposed ml-HCRF+ approach for (a) the Parliament \cite{VrigkasM_SETN14},
    (b) the TVHI \cite{Perez12}, (c) the SBU
    \cite{Ykiwon12}, and (d) the USAA \cite{FuHXG12} datasets using the CNN features.}
\label{fig:figure7}
\end{figure}                 
The corresponding confusion matrices of the proposed method for
all datasets, using the CNN-based features, are depicted in
Fig. \ref{fig:figure7}. The combination of privileged
information with the feature representation learned from the CNN model
resulted in very small inter- and
intra-class classification errors for all datasets.

\section{Discussion}
\label{sec:discussion}
Our method is able to robustly use privileged
information in a more efficient way than the SVM+ and the other
LUPI-based methods, by exploiting the hidden dynamics between
the video clips and the privileged information. We may also
observe that the proposed method outperforms all methods that
do not incorporate privileged information during learning.
Since the combination of multimodal data falls natural to the
human perception of understanding complex activities, the
incorporation of such information constitutes a strong attribute
for discriminating between different classes, rather than
learning each modality separately.

\textbf{Statistical significance:} In order to provide a
statistical evidence of the recognition accuracy, we computed
the p-values of the obtained results with respect to the
compared methods. Results highlighted with \colorbox{blue!15}{light purple} in
Tables \ref{Tab:table3} - \ref{Tab:table7} indicate
statistically significant improvement (p-values were less than
the significance level of $0.05$) over the second best method
using paired t-test.

\textbf{Computational complexity:} The proposed method uses the
same sufficient statistics as HCRF and the computational
complexity is similar to HCRF. The complexity of our method is
determined by the complexity of the corresponding inference
problem and is quadratic to the number of hidden states.

\subsection{Why is Privileged Information Important?}
\label{subsec:valuablePI}
Selecting which features can act as privileged information is
not an easy task. The performance of LUPI-based classifiers
relies on the delicate relationship between the regular and the
privileged information. Also, privileged information is costly
or difficult to obtain with respect to producing additional
regular training examples \cite{SerraToro14}. In general, when
privileged information alone is used as regular it may not be
sufficient for the correct classification of an action into its
respective category, since finding proper privileged
information is not always a straightforward process.

The scope of our approach is not to achieve the best results
possible but to investigate to what extent privileged
information can be beneficial under the same evaluation
protocol. The main strength of the proposed method is that it
achieves good classification results, when the LUPI framework
is incorporated with the standard HCRF model.

In the era of deep learning, significant progress has been
made in learning good representations of the data and a deep
learning based technique is the way to go. However, in cases
where datasets are small in size, which is true in our case, and the
distribution of the data is completely different from the data
that the existing pre-trained models were trained on, then
privileged information can be very helpful. Nonethenless, one
may fine-tune the deep neural model and extract meaningfull 
feature representations. This enhances our choice to use deep 
features with the proposed HCRF+ model as the experimetal results 
indicate significant improvement when these features are used.
Thus, the answer to the question ``is privileged information necessary?'' is
affirmative. For example, in many medical applications, where annotated
data are difficult or expensive to obtain and pre-trained deep
learning models are still not available, privileged information
is the best solution to go.

\section{Conclusion}
\label{sec:conclusions}

In this paper, we addressed the problem of human activity
categorization in a supervised framework and proposed a novel
probabilistic classification model based on robust learning
using a privileged information paradigm, called HCRF+. Our
model is made robust using Student's \textit{t}-distributions
to model the conditional distribution of the privileged
information. We proposed two variants for training in the LUPI
framework. The first variant uses maximum likelihood and the
second uses maximum margin learning.

Using auxiliary information about the input data, we were able
to produce better classification results than the standard HCRF
\cite{Quattoni07} approach. We evaluated the performance of our
method on four publicly available datasets and tested various
forms of privileged information. The experimental results
indicated that robust privileged information along with the
regular input data for training the model ameliorates the
recognition performance. We demonstrated improved results with
respect to the state-of-the-art LUPI framework especially when
CNN features are employed. 

According to our results, the proposed method and its variants
achieved notably higher performance than the majority of the
compared classification schemes. We were able to flexibly
understand multimodal human activities with high accuracy, when
not the same amount of information is available during testing.
By automatically estimating the regularization parameters
during learning, we managed to achieve high recognition
accuracy with less effort than standard cross validation based
classification schemes.

\appendices
\section{Conditional Distribution of the Privileged Information}
\label{AppendixA}

Recall that $\mathbf{x} \in \mathbb{R}^{M_{\mathbf{x}} \times T}$ is
an observation sequence of length $T$ and $\mathbf{x}^{*} \in
\mathbb{R}^{M_{\mathbf{x}^{*}} \times T}$ corresponds to the
privileged information of the same length. We partition the original
set $\left(\mathbf{x}^{*}, \mathbf{x}\right)^{T} \in \mathbb{R}^{M
\times T}$ into two disjoint subsets, where $\mathbf{x}^{*}$ forms
the first $M_{{\mathbf{x}}^{*}}$ components of $\left(\mathbf{x}^{*},
\mathbf{x}\right)^{T} \in \mathbb{R}^{M \times T}$ and $\mathbf{x}$
comprises the remaining $M - M_{\mathbf{x}}$ components. If the joint
distribution $p(\mathbf{x},\mathbf{x}^{*};\mathbf{w})$ follows a
Student's \textit{t}-law, with mean vector
$\mu=\left(\mu_{\mathbf{x}^{*}}, \mu_{\mathbf{x}}\right)^{T}$, a
real, positive definite, and symmetric $M \times M$ covariance matrix
$\Sigma =
\begin{pmatrix}
\Sigma_{\mathbf{x}^{*}\mathbf{x}^{*}} & \Sigma_{\mathbf{x}^{*}\mathbf{x}} \\
\!\!\Sigma_{\mathbf{x}\mathbf{x}^{*}} &
\!\!\Sigma_{\mathbf{x}\mathbf{x}}
\end{pmatrix}$ and $\nu \in [0,
\infty)$ corresponds to the degrees of freedom of the distribution
\cite{Kotz04}, then the conditional distribution
$p(\mathbf{x}|\mathbf{x}^{*};\mathbf{w})$ is also a Student's
\textit{t}-distribution:

\begin{equation}
\label{eq:Conditional.1}
\begin{split}
p(\mathbf{x}^{*}|\mathbf{x};\mathbf{w}) &= \text{St}(\mathbf{x}^{*};\mu^{*},\Sigma^{*},\nu^{*}) \\
& = \frac{ \Gamma\left(\left(\nu^{*}+M\right)/2\right) |\Sigma_{\mathbf{x}\mathbf{x}}|^{1/2} }
{ \left(\pi\nu^{*}\right)^{M_{\mathbf{x}}/2}
\Gamma\left(\left(\nu^{*}+M_{\mathbf{x}}\right)/2\right) |\Sigma^{*}|^{1/2}  } \\
& \quad\quad \times
\frac{ \left[1+\frac{1}{\nu^{*}} \mathbf{x}^{T}\Sigma_{\mathbf{x}\mathbf{x}}^{-1}\mathbf{x}
\right]^\frac{\left(\nu^{*}+M_{\mathbf{x}}\right)}{2} }
{\left[1+\frac{1}{\nu^{*}} Z^{T}{\Sigma^{*}}^{-1}Z \right] ^\frac{\left(\nu^{*}+M\right)}{2}} \, .
\end{split}
\end{equation}
The mean $\mu^{*}$, the covariance matrix $\Sigma^{*}$ and the
degrees of freedom $\nu^{*}$ of the conditional distribution
$p(\mathbf{x}^{*}|\mathbf{x};\mathbf{w})$, are computed by the
respective parts of $\mu$ and $\Sigma$:
\begin{align}
\label{eq:Conditional.2a}
\mu^{*} &= \mu_{\mathbf{x}^{*}} - \Sigma_{\mathbf{x}^{*}\mathbf{x}}
\Sigma_{\mathbf{x}\mathbf{x}}^{-1} \left(\mathbf{x} -
\mu_{\mathbf{x}}\right) \, ,
\\
\label{eq:Conditional.2b}
\Sigma^{*} &= \frac{\nu_{\mathbf{x}^{*}} + \left(\mathbf{x} -
\mu_{\mathbf{x}}\right)^{T}
\Sigma_{\mathbf{x}\mathbf{x}}^{-1}\left(\mathbf{x} -
\mu_{\mathbf{x}}\right)}{\nu_{\mathbf{x}^{*}} +
M_{{\mathbf{x}}^{*}}} \nonumber \\
& \quad\quad \times\left(\Sigma_{\mathbf{x}^{*}\mathbf{x}^{*}} -
\Sigma_{\mathbf{x}^{*}\mathbf{x}}\Sigma_{\mathbf{x}\mathbf{x}}^{-1}
\Sigma_{\mathbf{x}\mathbf{x}^{*}}\right)\, ,
\\
\label{eq:Conditional.2c}
\nu^{*} &= \nu_{\mathbf{x}^{*}} +
M_{{\mathbf{x}}^{*}} \, .
\end{align}
The parameters $(\mu,\Sigma,\nu)$ of the joint Student's
\textit{t}-distribution $p(\mathbf{x}^{*},\mathbf{x};\mathbf{w})$,
which are defined by the corresponding partition of the vector
$\left(\mathbf{x}^{*}, \mathbf{x}\right)^{T}$, are estimated using
the expectation-maximization (EM) algorithm \cite{Kotz04}. Then, the
parameters of the conditional distribution
$p(\mathbf{x}^{*}|\mathbf{x};\mathbf{w})$ are computed using Eq.
\eqref{eq:Conditional.2a}-\eqref{eq:Conditional.2c}.

It is worth noting that by letting the degrees of freedom $\nu^{*}$
to go to infinity, we can recover the Gaussian distribution with the
same parameters. If the data contain outliers, the degrees of freedom
parameter $\nu^{*}$ are weak and the mean and covariance of the data
are appropriately weighted in order not to take into account the
outliers.


\ifCLASSOPTIONcompsoc
  \section*{Acknowledgments}
\else
  \section*{Acknowledgment}
\fi

This work has been funded in part by the UH Hugh Roy and
Lillie Cranz Cullen Endowment Fund and the European Commission (H2020-MSCA-IF-2014),
under grant agreement No 656094. All statements of fact, opinion
or conclusions contained herein are those of the authors and should
not be construed as representing the official views or policies of
the sponsors.

\bibliographystyle{IEEEtran}
\bibliography{Vrigkas-LUPIHCRF-arxiv-bib-v01}

\end{document}